\documentclass[a4paper,fleqn]{cas-dc}

\usepackage[numbers]{natbib}
\usepackage{xcolor}
\newcommand{\red}[1]{\textcolor{red}{#1}}
\newcommand{\etal}[1]{\textit{et al.}{#1}}
\newcommand{\eg}[1]{\textit{e.g.}{#1}}

\def\tsc#1{\csdef{#1}{\textsc{\lowercase{#1}}\xspace}}
\tsc{WGM}
\tsc{QE}
\begin{document}
\let\WriteBookmarks\relax
\def\floatpagepagefraction{1}
\def\textpagefraction{.001}

\shorttitle{MCW-Net: Single Image Deraining with Multi-level Connections and Wide Regional Non-local Blocks}    

\shortauthors{Y. Park, M. Jeon, J. Lee, M. Kang}  

\title[mode = title]{MCW-Net: Single Image Deraining with Multi-level Connections and Wide Regional Non-local Blocks}  

\tnotemark[1] 

\tnotetext[1]{} 


\author[1]{Yeachan Park}[orcid=0000-0002-4211-6226]
\fnmark[1]

\ead{ychpark@snu.ac.kr}

% \ead[url]{<URL>}

\credit{Conceptualization, Methodology, Software, Writing-Original Draft, Writing - Review $\&$ Editing, Investigation, Formal analysis}

\affiliation[1]{organization={Department of Mathematical Sciences, Seoul National University},
            addressline={1 Gwanak-ro, Gwanak-gu}, 
            city={Seoul},
%          citysep={}, % Uncomment if no comma needed between city and postcode
            postcode={08826}, 
            country={Korea}}

\author[2]{Myeongho Jeon}[orcid=0000-0002-4509-582X]

% Footnote of the second author
\fnmark[1]

% Email id of the second author
\ead{andyjeon@snu.ac.kr}


\credit{Conceptualization, Software, Visualization, Investigation, Writing-Original Draft, Writing - Review $\&$ Editing, Data Curating, Formal analysis}


\affiliation[1]{organization={Department of
Computational Science and Technology, Seoul National University},
            addressline={1 Gwanak-ro, Gwanak-gu}, 
            city={Seoul},
%          citysep={}, % Uncomment if no comma needed between city and postcode
            postcode={08826}, 
            state={},
            country={Korea}}

\author[2]{Junho Lee}[orcid=0000-0001-7643-1024]

% Footnote of the second author
\fnmark[1]
% Email id of the second author
\ead{joon2003@snu.ac.kr}
\credit{Methodology, Software, Visualization, Investigation, Writing-Original Draft, Writing - Review $\&$ Editing, Data Curation, Validation}
% \affiliation[3]{organization={Department of
% Computational Science and Technology, Seoul National University},
%             addressline={1 Gwanak-ro, Gwanak-gu}, 
%             city={Seoul},
% %          citysep={}, % Uncomment if no comma needed between city and postcode
%             postcode={08826}, 
%             country={Korea}}

\author[1]{Myungjoo Kang}[]
\cormark[1]
% Footnote of the second author
% \fnmark[4]
% Email id of the second author
\ead{mkang@snu.ac.kr}
\ead[url]{ncia.snu.ac.kr}
\credit{Supervision, Writing-Review $\&$ Editing, Resources, Funding acquisition, Project administration}
% \affiliation[4]{organization={Seoul National University},
%             addressline={1 Gwanak-ro, Gwanak-gu}, 
%             city={Seoul},
% %          citysep={}, % Uncomment if no comma needed between city and postcode
%             postcode={08826}, 
%             country={Korea}}
% % Corresponding author text

\cortext[1]{corresponding author}

\fntext[1]{Equal Contribution}


\begin{abstract}
% \textcolor{red}{Removing rain streaks from single images is an important problem in various computer vision tasks because rain streaks can degrade outdoor images and reduce their visibility.} 
A recent line of convolutional neural network-based works has succeeded in capturing rain streaks. However, difficulties in detailed recovery still remain. In this paper, we present a multi-level connection and wide regional non-local block network (MCW-Net) to properly restore the original background textures in rainy images. Unlike existing encoder-decoder-based image deraining models that improve performance with additional branches, MCW-Net improves performance by maximizing information utilization without additional branches through the following two proposed methods. The first method is a multi-level connection that repeatedly connects multi-level features of the encoder network to the decoder network. Multi-level connection encourages the decoding process to use the feature information of all levels. In multi-level connection, channel-wise attention is considered to learn which level of features is important in the decoding process of the current level. The second method is a wide regional non-local block. As rain streaks primarily exhibit a vertical distribution, we divide the grid of the image into horizontally-wide patches and apply a non-local operation to each region to explore the rich rain-free background information. Experimental results on both synthetic and real-world rainy datasets demonstrate that the proposed model significantly outperforms existing state-of-the-art models. Furthermore, the results of the joint deraining and segmentation experiment prove that our model contributes effectively to other vision tasks. 
\end{abstract}


% \begin{highlights}
% \item Propose multi-level connection (MLC) between levels to restore the details of the rainy image.
% \item Propose a wide regional non-local block providing a relatively even distribution of the rain streaks by region
% \item Show efficient and state-of-the-art performance in rain removal. 
% \item Contribute to other vision tasks by enhancing visibility under adverse weather conditions.
% \end{highlights}

% Keywords

\maketitle

% Main text
\section{Introduction}
\label{sec:intro}

Adverse weather conditions such as rain, haze, and snow can produce complex visual effects on natural images and videos. In particular, rain streaks, which is one of the most commonly occurring phenomena in outdoor imaging, can potentially degrade the performance in several computer vision applications. Therefore, it is imperative to develop algorithms that effectively remove rain streaks and restore pristine background scenes in vision-related tasks.

Over the past few decades, several research works have studied the removal of rain streaks from captured images. Several traditional deraining methods have suggested separating rain streaks from the clean background image based on the physical characteristics or texture appearance patterns of the rain streaks. Recently, convolutional neural network (CNN)-based methods have achieved great success in solving this problem \cite{jiang2020multi,li2018non,li2018recurrent,ren2019progressive,wang2019erl,wang2019spatial,yang2019scale,yang2017deep,yu2019gradual,zhang2019image}.

%traditional methods
%\cite{barnum2010analysis,bossu2011rain,chen2013generalized,kang2011automatic,li2016rain,luo2015removing}

% \textcolor{blue}{Many of the CNN-based methods have encoder-decoder structures, and for the most part, they sought to improve performance by adding subnetworks without fully utilizing the information generated during the encoding-decoding process. For example, to remove fine-grained rain streaks and recover rain-free backgrounds more clearly, Yu \textit{et al.} \cite{yu2019gradual} proposed a two-stage model using encoder-decoder as a coarse deraining stage and a simple network as a fine deraining stage. And Wang \textit{et al.} \cite{wang2019erl} added a residual learning branch parallel to the encoder part to form a better conditional embedding and eventually generate a much better deraining result in the decoder part. Adding these subnetworks improves performance, but there is a limitation that the model simply becomes heavy without leveraging enough information.}

Many of the CNN-based methods utilize encoder-decoder structures, and for the most part, they add subnetworks without fully utilizing the information generated during the encoding-decoding process. For example, to remove fine-grained rain streaks and recover rain-free backgrounds more clearly, Yu \textit{et al.} \cite{yu2019gradual} consider the encoder-decoder as a coarse deraining stage and use an additional simple network as a fine deraining stage. Wang \textit{et al.} \cite{wang2019erl} add a residual learning branch parallel to the encoder part to form a better conditional embedding and eventually generate a much better deraining result in the decoder part. Adding these subnetworks can easily improve performance, but there is a limitation that a model becomes heavier without leveraging enough information of an original model.

There is also an effort to utilize the information that is generated within the model. In order to obtain and leverage information from other pixels for the degraded background pixels, Li \textit{et al.} \cite{li2018non} and Yu \textit{et al.} \cite{yu2019gradual} exploit non-local operations. These models use a square grid with the same aspect ratio in non-local operations. However, the operations with the square grid lack an understanding of the unique properties of the rain streaks because of their vertical distribution in the rainy image, which we explore (see Figure \ref{fig:rain distribution}). Consequently, these methods have difficulties in recovering details in extremely adverse weather conditions.

\begin{figure*}[!t]
	\centering\includegraphics[width=0.9\linewidth]{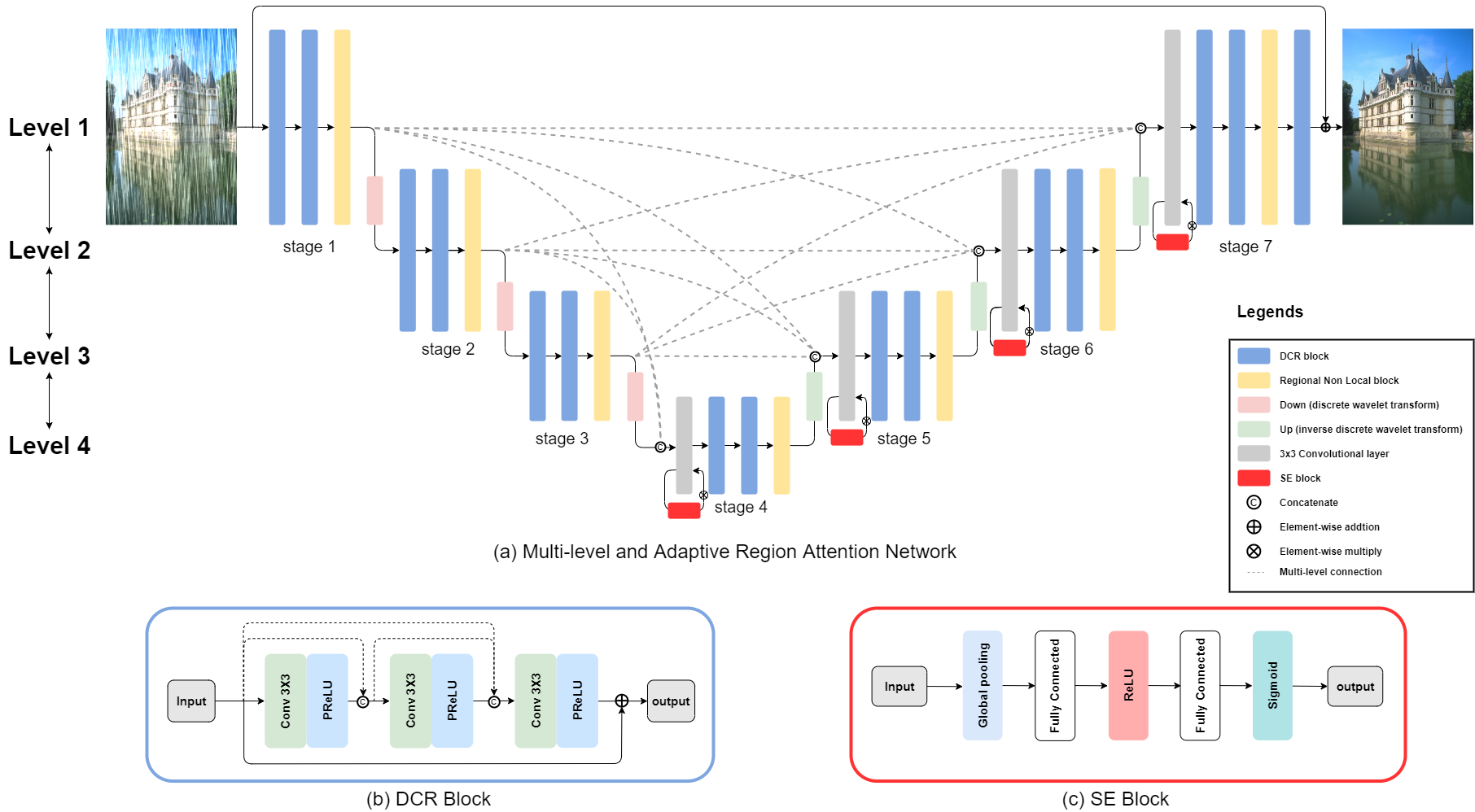}\\
	\caption{Overview of the proposed MCW-Net structure.}

	\label{fig:overview}
\end{figure*}

To address these limitations of the prior works, we present a multi-level connection and wide regional non-local block network (MCW-Net) to carefully remove rain streaks and recover background details efficiently leveraging information generated during the encoding-decoding process.
The proposed MCW-Net is based on an encoder-decoder structure consisting of down-sampling and up-sampling components as depicted in Figure \ref{fig:overview}.

We construct multi-level connection (MLC) between multiple-scale features to efficiently utilize information across various scales without additional subnetworks in the recovery of the background. We implement an interactive multi-connection that considers the interconnections between different scales. Because the features at multiple levels show different scale characteristics, direct connections rather cause adverse effects in the model. To adaptively rescale the channel-wise features in MLC, we apply a channel-wise attention layer \cite{hu2018squeeze} after MLC, which helps the network to focus on the useful channels. We demonstrate the importance of the channel-wise attention, and we validate that MLC plays an effective role by comparing the qualitative and quantitative results of models with and without MLC in Section \ref{sec:mlc_ablation}.
% Table \ref{tab:mlc-ablation} shows the importance of channel-wise attention, and we validate that multi-scale connection plays an effective role by comparing the qualitative and quantitative results of models with and without multi-scale connection in Table \ref{total ablation} and Figure \ref{fig:mlc_ablation}.

In addition, we implement a non-local operation \cite{NonLocal2018} to capture long-range spatial dependencies between distant pixels. We propose a wide regional non-local block (WRNL), which divides feature maps into grids of wide regions (see Figure \ref{fig:nonlocal_block}) before performing the region-wise non-local operation. This wide grid provides a relatively more even distribution of the rain streaks by region, which facilitates the retrieval of rich long-range background information during the recovery of the original rain-free image (see Section. \ref{analysis}).

Additionally, as described in \cite{yang2019scale}, to prevent information loss during the sampling operation, we adopt the discrete wavelet transform (DWT) and inverse DWT (IWT) in place of the simple pooling and de-convolution operations. Unlike the pooling operation, the DWT operation is invertible via IWT, which helps to avoid information loss. In addition, rain streaks can be captured with rich frequency information via wavelet transform.
 
We evaluate the proposed MCW-Net on various synthetic and real-world deraining datasets and compare its performance with existing state-of-the-art methods. 
In particular, for real-world images, we measure the performance of the proposed method using B-FEN \cite{wu2020subjective} metric dedicated to deraining quality measurement. We conduct an experiment on raindrop data, another degradation phenomenon caused by rain from the perspective of the generalization ability of the model. In addition, we validate in RainCityscape experiments that the proposed method can also help with other vision tasks such as semantic segmentation.
\\ 
In summary, the contributions of this work may be summarized as follows.
\noindent  \textbf{1)}
We propose MLC to fully leverage information generated in encoding-decoding process for detail recovery without additional subnetworks. Feature information of all the scales in the down-sampling part is aggregated at each stage of the up-sampling part of the network, so it helps to recover details by preventing information loss that occurs during the sampling process. We also analyze that channel-wise attention plays an key role in the MLC.

\textbf{2)}
We propose the WRNL, which effectively restores the background by using sufficient rain-free information in each region of widely divided grids in the input feature maps. We experimentally demonstrated that the distribution of even rain streaks by grid helps the deraining performance.

\textbf{3)}
We perform experiments on both synthetic and real-world rain datasets and demonstrate that the proposed method significantly outperforms existing state-of-the-art methods. We also demonstrate the excellence of the proposed method for real-world images using B-FEN, a metric dedicated to measuring deraining quality.

\textbf{4)}
We construct joint image deraining and semantic segmentation models on the RainCityscape dataset. In addition to conventional comparisons such as the peak signal-to-noise ratio (PSNR) and structural similarity index measure (SSIM), we comprehensively evaluate the contribution of the deraining model to other vision tasks.

\section{Related Work}

The single image deraining problem begins with the assumption that a rainy image consists of a background layer and a rainy layer. Several traditional training methods based on single images and videos have been proposed. Barnum \textit{et al.} \cite{barnum2010analysis} reconstruct rainy images by combining the appearance model with the streak model. The appearance model identifies individual rain streaks and the streak model utilizes the statistical characteristics of rain. Chen and Hsu \cite{chen2013generalized} use the low-rank model to separate the layers in a rainy image. As noted by Yang \textit{et al.} \cite{yang2019single}, sparse coding is applied during this process to separate the rainy layer from the rainy image \cite{deng2018directional, kang2011automatic, luo2015removing, wang2017hierarchical, zhu2017joint}. Further, Li \textit{et al.} \cite{bossu2011rain, li2016rain} approach this problem using the Gaussian mixture model.

 Because of the remarkable performance exhibited by deep learning-based methods, especially CNN-based ones, the potential use of deep learning in deraining has been extensively researched. Yang \textit{et al.} \cite{yang2017deep} apply a CNN-based method for the first time and express natural images by adding atmospheric light as a component to rainy images. Fu \textit{et al.} \cite{fu2017removing} and Fan \textit{et al.} \cite{fan2018residual} use a single primary network that restores input images using the residual network. Based on the residual network, Li \textit{et al.} \cite{li2018recurrent} attempt to further eliminate overlapping rain streaks by organizing the context aggregate network into multiple stages. Shen \textit{et al.} \cite{shen2018deep} consider rain streaks to be high-frequency and attempt to remove rain streaks by utilizing DWT. Yang \textit{et al.} \cite{yang2019scale} divide the deraining process into several stages and reconstruct the image recurrently, beginning with a small portion of the image to eventually obtain the entire image. 
 
Wang \textit{et al.} \cite{wang2019spatial} capture the spatial contextual information using a four-directional recurrent neural network with the identity matrix initialization model. Ren \textit{et al.} \cite{ren2019progressive} propose progressive ResNet to effectively remove the rain via recursive computation. Yu \textit{et al.} \cite{yu2019gradual} propose GraNet, which is designed to identify rain masks in the coarse stage using a region-aware non-local block. Subsequently, the process uses the rain masks to create the final image using another reconstruction network. To achieve pixel-wise deraining in image recovery, encoder-decoder structures have been used in certain methods. Wang \textit{et al.} \cite{wang2019erl} propose the residual learning branch as a component of the encoder. Li \textit{et al.} \cite{ li2018non} enhance the performance by introducing non-local blocks into the encoder-decoder network. Among the methods that reconstruct the rainy layer to be identical to the background layer, the generative adversarial network is widely used to remove raindrops and rain streaks \cite{ li2019heavy, qian2018attentive, zhang2019image}.

Yang \textit{et al.} \cite{yang2020towards} propose the fractal band learning network based on frequent band recovery. Wang \textit{et al.} \cite{wang2020model} propose an interpretable deep network based on a convolutional dictionary network. Jiang \textit{et al.} \cite{jiang2020multi} use the images of various sizes as the input to the model. A multi-scale pyramid structure is used to promote cooperative representation. Deng \textit{et al.} \cite{deng2020detail} propose two-branch parallel networks, in which one branch performs rain removal and the other branch detail recovery. In \cite{wang2020rethinking}, newly formulated rain streaks transmission maps, vapor transmission maps, and atmospheric lights are respectively learned by three different networks. Zhang \textit{et al.} \cite{zhang2020beyond} propose a paired rain removal network, which exploits both stereo images and semantic information. Zamir \textit{et al.} \cite{zamir2021multi} propose a multi-stage progressive architecture with a supervised attention module for image restoration.

Chen \etal~\cite{chen2021pre} present an image processing transformer (IPT). IPT covers different several tasks such as super-resolution, denoising, and deraining based on the transformer method. The authors augment ImageNet images to low-resolution, noised, and rainy images via corresponding filters and then pre-train the IPT with each set. Yue \etal~\cite{yue2021semi} propose a dynamic rain generator to mimic the rain streaks in the video. The rain streaks in generated videos are removed by a deep learning-based model called derainer.

Zhang \etal~\cite{zhang2020multi} exploit the low to high-level features and attention operation to restore the hazed images. Their intuition is that the low-level features contribute to recovering finer details and the high-level features represent the shape of the object or abstract semantic information. The utilization of hierarchical features and the attention mechanism are similar to one of our strategies, multi-level connection. However, their work fuses the lower-level features only in the most-down-sampled features, whereas we consider all the features captured in the down-sampling phase on every up-sampling phase.
%chen2021pre, yue2021semi, zhang2020multi

\section{Proposed Network}
\label{sec:proposed}

In this section, we describe the architecture of the proposed MCW-Net, which is is based on a U-Net-like structure whose overview is depicted in Figure \ref{fig:overview}. As is apparent from the figure, we divide the levels according to the size of the feature map and define a set of blocks as a stage.

 The proposed MCW-Net consists of an encoder part and a decoder part. The first three stages form the encoder part, and the remaining four stages the decoder part. We propose MLC, which connect all outputs of the encoder to all inputs of the decoder. MLC enables more diverse scale features to be used during the restoration process. Each stage of MCW-Net is composed of two densely connected residual (DCR) blocks, each of which consists of three convolution layers followed by PReLU \cite{trottier2017parametric} (refer Figure \ref{fig:overview}(b)) and one WRNL block. To adaptively rescale channel-wise features after concatenating the multi-level features, a squeeze-and-excitation (SE) block is added in front of each decoder stage. A 1$\times$1 convolutional layer follows the SE block to adjust the number of channels. 

\subsection{Multi-Level Connection}
\label{sec:multi-level connections}

% \indent\textbf{Squeeze-and-Excitation Block}
% \label{subsec:Squeeze-and-Excitation Block}
% The idea of channel-wise attention has been widely used in CNNs to utilize channel-wise dependencies between features. The SE network \cite{hu2018squeeze} is adopted in the proposed network to rescale channel-wise features. In SENet, channel-wise statistics are obtained via global average pooling (GAP) to exploit global contextual information lying outside the local receptive regions. To fully capture channel-wise dependencies, a gating mechanism is introduced to form two fully-connected (FC) layers that reduce or increase the dimensionality with the ratio $r$. Following \cite{zhang2018image}, we incorporate the idea of the residual block \cite{lim2017enhanced} with channel-wise attention (see Figure \ref{fig:overview}) to construct the SE block. 

% Formally, given an input feature map $X$, the SE Block $f_{SE}(\cdot)$ can be expressed as follows:  
% \begin{align}
%  &S        =  \sigma( W_{2} \delta ( W_{1} H_{GAP}(X) ) ) \\
%  &f_{SE}(X) = X+ \hat{X} = X + S \otimes_{c} X 
% \end{align}
% where $\otimes_{c} $ denotes channel-wise multiplication, $H_{GAP}(\cdot)$ denotes the global average pooling function, $\sigma$ denotes the sigmoid function, $\delta$ denotes the ReLU function, $W_{1} \in \mathbb{R}^{\frac{C}{r}\times C}$ and $W_{2} \in \mathbb{R}^{ C \times \frac{C}{r}}$ denote the dimension reduction and dimension-increasing FC layers respectively, and $r$ denotes the reduction ratio in the gating mechanism.

% \indent\textbf{Multi-Level Connections}
In the usual U-Net-like network, connections exist only between features corresponding to the same level. Such a structure cannot make use of multiple scale information during the recovery of low-level features in the decoder. However, single image deraining is a low-level vision task that requires richer range scale features to restore the details in the image. Inspired by \cite{sun2019highresolutionposeestimation, tan2019efficientdet, wang2019highresolutionvisualrecognition}, we formulate MLC to aggregate the features of all the levels. At each stage of the up-sampling part of the network, features from all scales in the down-sampling part is aggregated. These multi-scale features provide a wider range of information from simple patterns (\textit{e.g.}, corners or edge/color conjunctions) in its lower-level to more complex high-level features (\textit{e.g.}, significant variation and object-specific features). They encourage more delicate deraining because rainy pixels in the image are recovered referencing semantic context and details from other intact pixels. However, simply fusing various features might cause 
necessary information weighted insufficiently in conjunction with more weight on less helpful information at the current up-sampling stage. The attention mechanism allows the model to focus more on significant channels among several channels, and for this reason, it is essential when connecting features of multiple levels.
% For this reason, the attention mechanism is favorable when connecting multiple features.}
%  Because the features of multiple levels have different scale characteristics, we adopt the SE block to adaptively rescale the channel-wise features. 

Formally, let $ E_{out}^{l}$ be the output features at level $l$ $(l=1,2,3)$ in the encoder part. At each level $l$  $(l=1,2,3,4)$ in the decoder part, the input feature $D_{in}^{l}$ is given as: 
\begin{align}\label{eq:MLC}
  D_{concat}^{l} &= (\bigoplus_{i=1}^{3} H_{i}^{l} ( E_{out}^{i} ) )\oplus H_{up}(D_{out}^{l+1})  \\
  D_{in}^{l} &= W_{1\times1} ( f_{SE} ( D_{concat}^{l}  )  )
\end{align}
where $\oplus$ denotes the concatenation operation, $H_{up}(\cdot)$ denotes the up-sampling operation, $D_{out}^{l}$ denotes the output feature of the decoder part at level $l$, $ W_{1\times1}$ denotes the $1\times1$ convolution layer, and $f_{SE}(\cdot)$ denotes the SE block discussed above. $H_{i}^{l}(\cdot)$ denotes the sampling operation from level $i$ to $l$. In other words, $H_{i}^{l}$ is the down-sampling by $l-i$ times, identity, and up-sampling by $i-l$ times operations if $l>i$ , $l=i$, and $l<i$, respectively. We set $ D_{in}^{5} = 0 $  for convenience.

 Without MLC, high-level features cannot be used during the processing of low-level features and vice versa. This approach helps the network to exploit various scale representations in recovering large-scale features. To find the correct correspondence between the feature shapes at different scales, we apply discrete wavelet transforms (DWT or IWT), as described in Section \ref{sec:wavelet transform}, for the down-sampling and up-sampling operations.

\begin{figure*}[!t]\footnotesize
	\centering
	\setlength{\tabcolsep}{0pt}
	
	\begin{tabular}{cclcclccl}
		\multicolumn{3}{c}{\includegraphics[width=.4\textwidth]{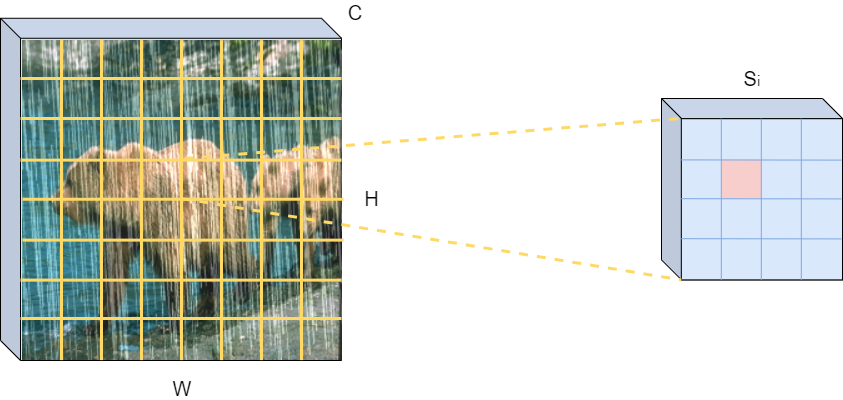}}\hspace{10mm}\ &
		\multicolumn{3}{c}{\includegraphics[width=.4\textwidth]{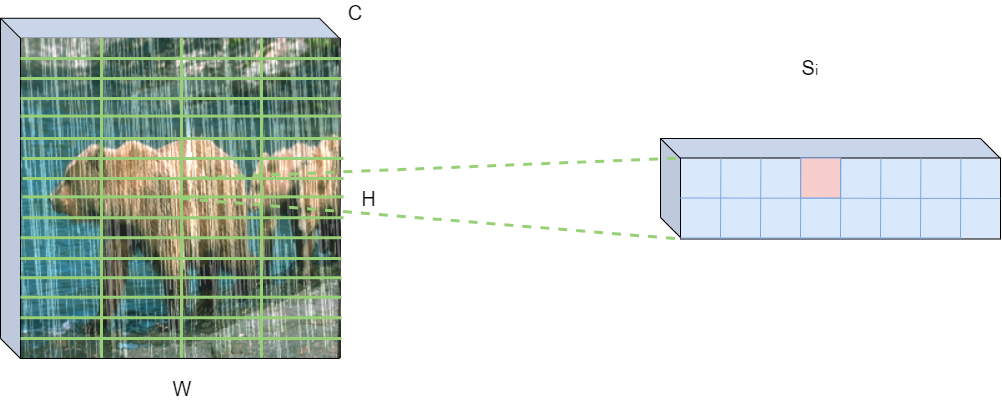}}\\
		\multicolumn{3}{c}{(a)} &
		\multicolumn{3}{c}{(b)} 
	\end{tabular}
	\caption{Examples of patches of the input feature of the regional non-local block. (a) Square patch, (b) Wide rectangular patch. Every pixel in a patch refers to every pixel in the patch. } 
	\label{fig:nonlocal_block}
\end{figure*}

\begin{figure*}[!t]\footnotesize
	\centering
	\setlength{\tabcolsep}{0pt}
	
	\begin{tabular}{cclcclccl}
		\multicolumn{3}{c}{\includegraphics[width=.33\textwidth]{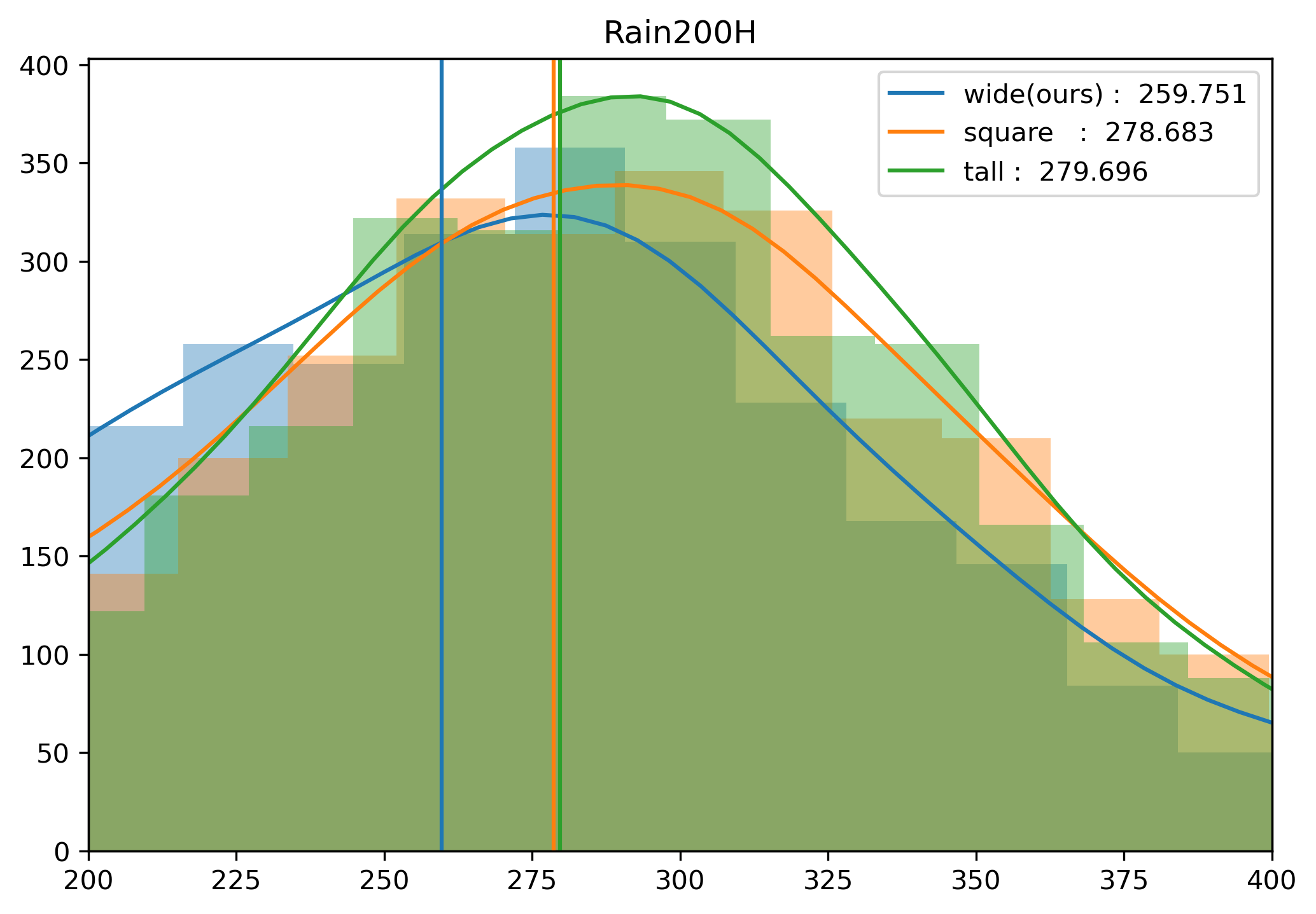}} &
		\multicolumn{3}{c}{\includegraphics[width=.33\textwidth]{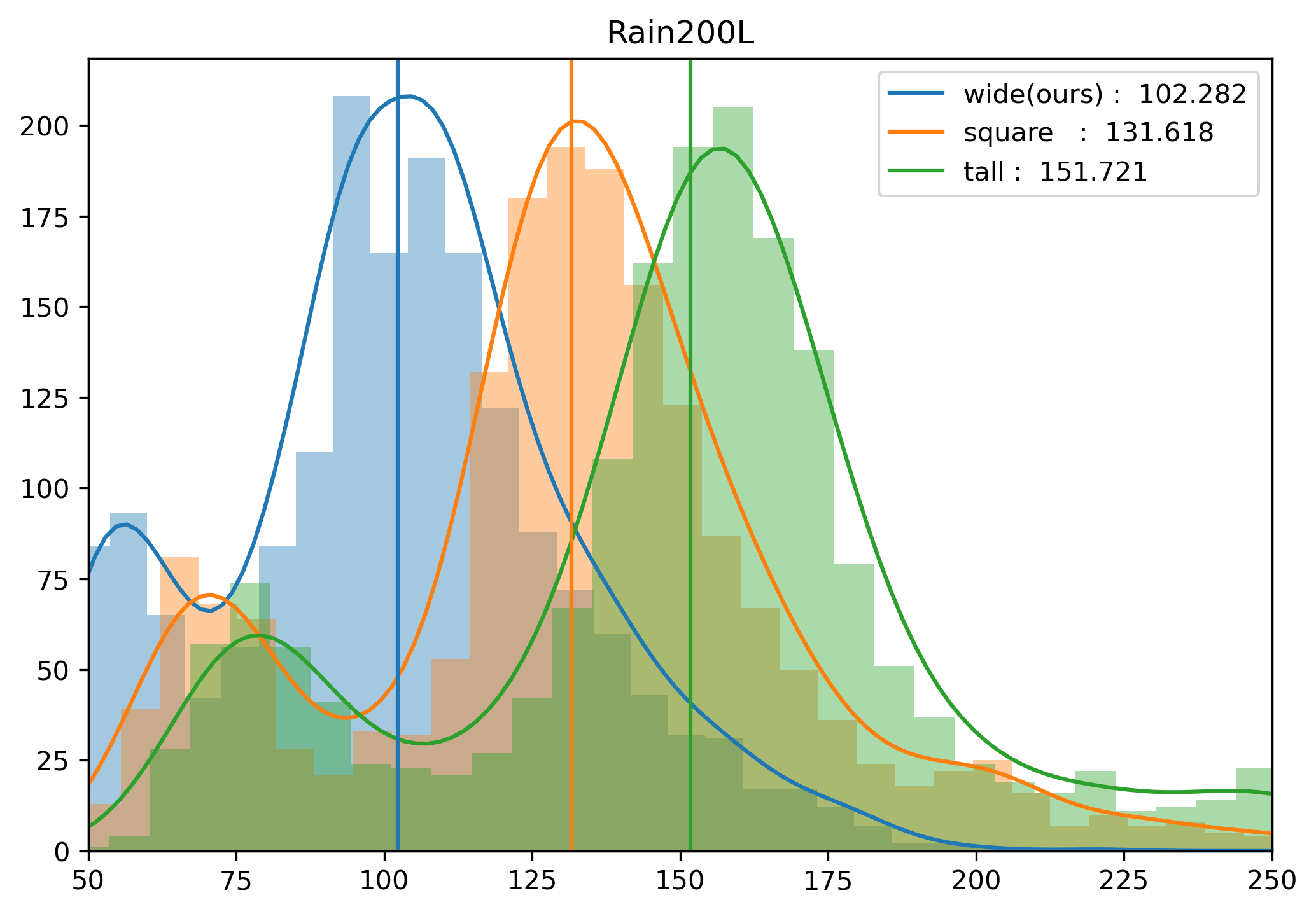}} &
		\multicolumn{3}{c}{\includegraphics[width=.33\textwidth]{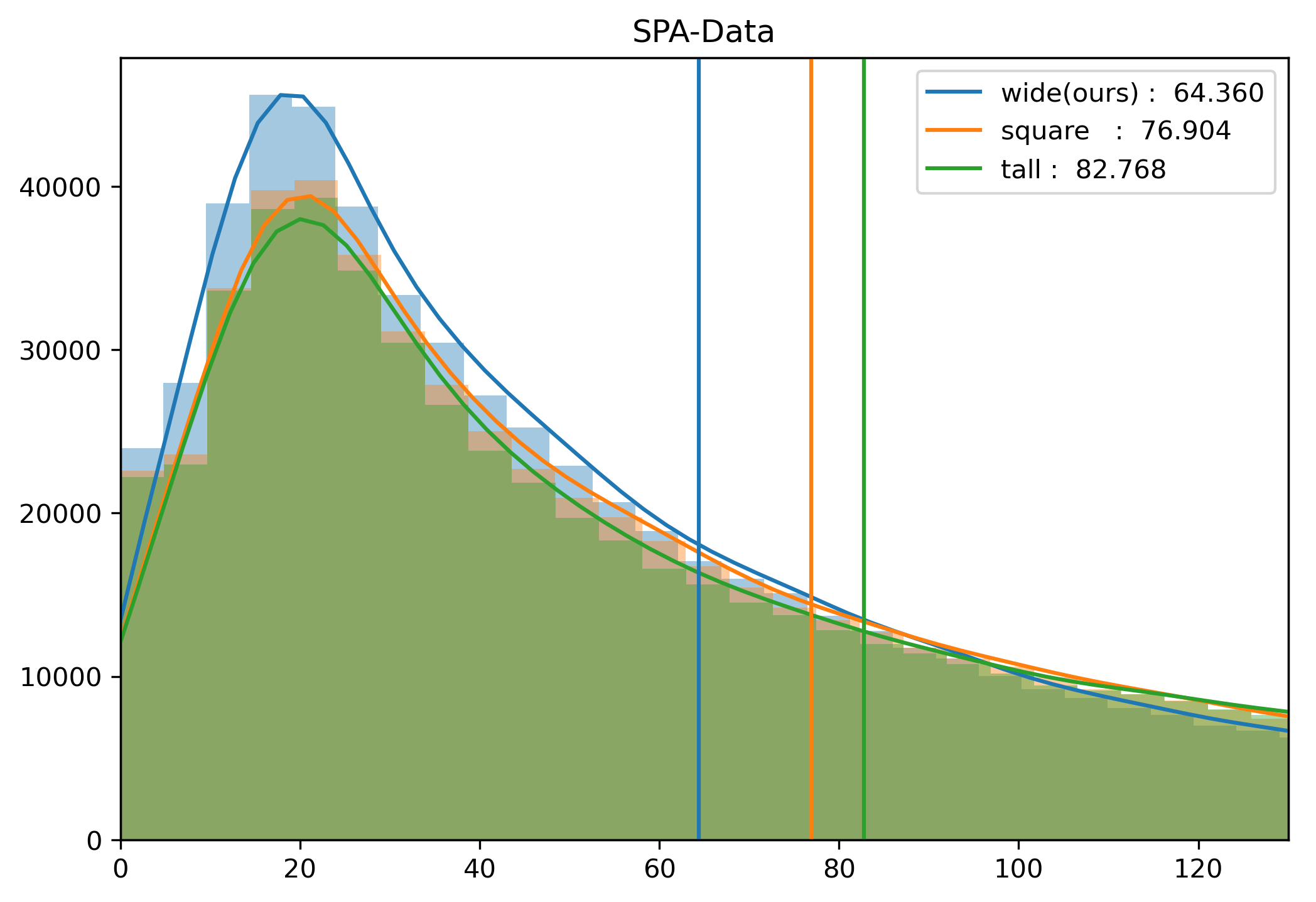}}
		\\
		\multicolumn{3}{c}{(a) Rain200H \cite{yang2017deep}} &
		\multicolumn{3}{c}{(b) Rain200L \cite{yang2017deep}} &
		\multicolumn{3}{c}{(c) SPA-data \cite{wang2019spatial}}
	\end{tabular}
	\caption{Analysis of rain streak distributions in various region types. The x-axis represents the standard deviation between the number of rain pixels in the patches in each image. The y-axis represents the number of images and vertical bars are the means of each dataset standard deviation. The distribution of the images according to the standard deviation is represented by histograms. We approximate the probability density function of the histogram by using kernel density estimation. As can be seen in the figures, the wide region has the smallest standard deviation mean on all the datasets, so it can be interpreted that each patch of the wide region has the evenest background information.}

	\label{fig:rain distribution}
\end{figure*}

\subsection{Wide Regional Non-Local Block}

% In this section, we first describe the representation of the WRNL block and then provide an analysis of the effectiveness of the WRNL block based on statistical exploration.

We denote the input feature to the WRNL as $ X  \in \mathbb{R}^{H \times W \times C}$. We divide $X$ into a $a \times b $ grid of patches $ \{ X^{k} \} ,( k=1, ..., K=ab ) $ where $K$ is the number of patches. The grid division is illustrated in Figure \ref{fig:nonlocal_block}. The linear embedding processes for $X^{k} $ to generate the output $Z^{k}$ are formulated as follows. 
\begin{align}  
\label{embedGaussian}
&\Phi({X^{k}})_i^j = \phi({X}_{i}^{k}, {X}_{j}^{k}) = \text{exp} \{ \theta({X}_{i}^{k}) \psi({X}_{j}^{k})^T \}  \\ % \,,\\
&\resizebox{.9\hsize}{!}{ $\theta({X}_{i}^{k})= {X}_{i}^{k} {W}_\theta, 
 \psi({X}_{i}^{k}) = {X}_{i}^{k} {W}_\psi, 
{G} ({X})_{i}^{k}  = {X}_{i}^{k} {W}_g  $  %\, ,\forall i\,. 
}
\end{align}

where $ {X}_{i}^{k}$ denotes the feature $X^{k}$ at position $i=1,...,HW/ab$. The learnable weight matrices ${W}_\theta$, ${W}_\phi$, and ${W}_g$ have the dimensions of $C \times L$, $C \times L$, and $ C \times C$, respectively. In practice,  $L = C/2$ is used. The regional non-local operation can be expressed as follows: 
\begin{align}  
\label{proposed}
{Z}_i^k = \frac{1}{\delta_i ({X^{k}})} \sum\nolimits_{j \in {S}_i} \Phi({X^{k}})_i^j \, {G} ({X^{k}})_i\,, \;\;\forall i\;,
\end{align}
where $\delta_i ({X^k}) = \sum\nolimits_{j \in {S}_i} \phi({X}_i^k, {X}_j^k)$ denotes the correlation between ${X}_i^k$ and each ${X}_j^k$ in ${S}_i$, and $Z_i^k$ denotes the output feature $Z^k$ at position $i$. ${S}_i$ denotes a set of patch positions. If $ a > b $, then the patch is wider than when $a=b$. Therefore, we call the patch a wide rectangular patch, a square patch, and a tall rectangular patch if $a > b$ , $a= b$, and $ a <b$, respectively. In the WRNL block, we set the $a \times b$ grids to $16 \times 4$ , $8 \times 2$, $4 \times 1$, and $4 \times 1$ at levels 1, 2, 3, and 4, respectively.

\subsubsection{Analysis}
\label{analysis}

Given that the non-local block recovers a specific pixel based on the information of other pixels in the patch, it is necessary to have sufficient background information in each patch. The regional non-local block uses the background information sufficiently if the rain streaks are evenly distributed between the patches. However, we observe that the rain streaks are not evenly distributed between square patches in the images used in the previous deraining research \cite{li2018non,yu2019gradual}. Because of the predominantly vertical distribution of rain steaks, we expect that wide rectangular patches have a more even distribution of the streaks than square and tall rectangular patches. 

% have small standard deviation values compared to square and tall rectangular patches, which means that rain is evenly distributed across all patches.

The distribution of the rain streaks is confirmed through experiments. Wide rectangular, square, and tall rectangular patches are prepared by dividing the height and width of the image into 16 $\times$ 4, 8 $\times$ 8 and 4 $\times$ 16 grids respectively. It should be noted that (a) in Figure \ref{fig:nonlocal_block} contains an 8 $\times$ 8 grid of patches, and (b) in Figure \ref{fig:nonlocal_block} contains 16 $\times$ 4 grid of patches. We consider pixels as rain streaks if the difference between the pixels in $x_{input}$ and $x_{gt}$ exceeds a certain threshold. The standard deviation between the number of rain pixels in the patches included in each image is depicted in Figure \ref{fig:rain distribution}. Wide rectangular patches are observed to exhibit smaller standard deviation values compared to square and tall rectangular patches, which implies an even distribution of rain across all patches. This results in the effective recovery of the image because the usable background information within each patch is also distributed evenly as shown in Table \ref{nl ablation}.

\subsection{Discrete Wavelet Transform}
\label{sec:wavelet transform}
To prevent information loss, we adopt the discrete wavelet transform for the sampling operation. In particular, We use 2D Haar wavelet which is widely used in image processing.

The proposed network uses DWT and IWT for down-sampling and up-sampling, respectively. In particular, we adopt the Haar transform, which is simple and widely used method in image processing \cite{guo2017deep,liu2018multi,porwik2004haar,shen2018deep,yang2019scale}. The Haar transform is calculated based on the filter $\mathbf{f}_{LL}$, $\mathbf{f}_{LH}$, $\mathbf{f}_{HL}$ and $\mathbf{f}_{HH}$ as follows: %search more haar transform for image transform 
\begin{equation}
\label{eq:haar_ll}
    \resizebox{.9\hsize}{!}{$
    \mathbf{f}_{LL} = {1 \over 4}\begin{bmatrix}
                    1 & 1\\
                    1 & 1
                    \end{bmatrix},
    \mathbf{f}_{LH} \!=\! {1 \over 4}\begin{bmatrix}
                    -1 \!\!&\!\! -1\\
                    1 \!\!&\!\! 1
                    \end{bmatrix},
    \mathbf{f}_{HL} \!=\! {1 \over 4}\begin{bmatrix}
                    -1 \!\!&\!\! 1\\
                    -1 \!\!&\!\! 1
                    \end{bmatrix},
    \mathbf{f}_{HH} \!=\! {1 \over 4}\begin{bmatrix}
                    1 \!\!&\!\! -1\\
                    -1 \!\!&\!\! 1
                    \end{bmatrix}.
                    $ }
\end{equation}

Given that $\mathbf{f}_{LL}$ is identical to average pooling, $LL$ achieves local translation invariance by reducing the size of the feature map (Equation \ref{eq:haar_ll}). $LH$, $HL$, and $HH$ contain edge information. In particular, as $LH$ contains vertical edge information, the features of the rain streaks can be effectively obtained from it. The IWT operation during the up-sampling process is the inverse operation of the DWT.

\subsection{Loss Function}
We define the loss function $L$ as follows. 
\begin{align}
% \footnotesize
% \mathcal {L} &= \mathcal{L}_{1} + \mathcal{L}_{2} 
\mathcal {L} &= \| x_{gt}-f (x_{input}) \| _{1} + \| x_{gt}-f (x_{input}) \| _{2}
% \mathcal{L}_{1} &= \| x_{gt}-f_{MCW-Net} (x_{input}) \| _{1} \\
% \mathcal{L}_{2} &= \| x_{gt}-f_{MCW-Net} (x_{input}) \| _{2}
\end{align}

where $x_{input}$ denotes the input rainy image, $x_{gt}$ denotes the corresponding rain-free image, and $f$ denotes the return of the MCW-Net output with respect to $x_{input}$. We use L1+L2 loss because it shows the slightly better performance, but our method does not appear to be sensitive to the loss.

\begin{table*}[]
	\tabcolsep 0.06in{\scriptsize{}}
	\centering
	\caption{Average PSNR and SSIM comparison on the synthetic datasets Rain200H \cite{yang2017deep}, Rain200L \cite{yang2017deep}, Rain800 \cite{zhang2019image}, Rain1200 \cite{zhang2018density}, and real-world dataset SPA-Data \cite{wang2019spatial}. The highest values are indicated in \textcolor{red}{red} and the second-highest values are indicated in \textcolor{blue}{blue}. 
% 	Note that although IPT use pre-trained weights trained with rainy-augmented ImageNet dataset for training, we manually train the IPT in the same dataset as that of the proposed method and all the other comparable models because rainy ImageNet is not provided.
	 Note that IPT uses rainy-augmented ImageNet pre-trained weight, but proposed methods and other comparable models do not. So we manually train the IPT from sketch only with the provided dataset.}
	\vspace{0.15cm}
	{\small{
    \begin{tabular*}{\tblwidth}{@{}LLLLLLL@{}}		
		
		\toprule
		Method   & JORDER \cite{yang2017deep} & RESCAN \cite{li2018recurrent}   & SPANet \cite{wang2019spatial}  & PReNet \cite{ren2019progressive}    &  ReHEN \cite{yang2019single}    &  RCDNet \cite{wang2020model} \\
	    & (CVPR' 2017)  & (ECCV' 2018)  & (CVPR' 2019) & (CVPR' 2019) & (MM' 2019) & (CVPR' 2020)  \\ 
		\midrule
		Rain200L  & 36.95/0.979                & 36.94/0.980                     & 35.60/0.974                    & 36.28/0.979                         & {38.57/0.983 }          & 35.28/0.971      \\ 
		Rain200H  & 22.05/0.727                & 26.62/0.841                     & 26.32/0.858                    & 27.64/0.884                         & 27.48/0.863                             & 26.18/0.835         \\ 
		Rain800    & 22.24/0.776                & 24.09/0.841                     & 24.37/0.861                    & 22.83/0.790                         &{26.96/0.854}        & 24.59/0.821       \\
		Rain1200  & 24.32/0.862                & 32.48/0.910                     & 32.38/0.920                    & 30.40/0.891                         & {32.64/0.914}       & 32.23/0.910       \\ 
% 		Rain1400 \cite{fu2017removing} & 27.55/0.853                & 28.57/0.891                     & 28.23/0.924                    & \textcolor{red}{32.60/0.946}                & 31.33/0.918                     & \textcolor{blue}{31.95/0.926}\\
		SPA-Data  & 35.72/0.978                & 36.99/0.967                     & 38.53/0.987                    & 35.68/0.942                         & 38.65/0.974          & {41.47/0.983}      \\ 
		Params & 4,169,024               & 499,668                     & 283,716                    & 168,963                         & 298,263          & 3,166,355      \\  % & 129,539,018
% 		Running Time & 0.00715               & 0.0172                     & 0.0077                    & 0.0290                        & 0.0009         & 0.09675       & -        & 0.1519 & 0.0893 \\
	
	\bottomrule
	
		Method      &  DRD-Net \cite{deng2020detail}  & MPRNet \cite{zamir2021multi} & IPT \cite{chen2021pre} & IPT \cite{chen2021pre}  & MCW-Net  & MCW-Net\\
	      &  (CVPR' 2020) & (CVPR' 2021) & (CVPR' 2021) & (w/ pretraining) & (Ours-small) & (Ours-large)\\ 
		\midrule
		Rain200L  &    37.15/0.987                      &37.87/0.983 & 37.08/0.980 &\textcolor{red}{40.32/0.989}& {39.19/0.986} & \textcolor{blue}{39.92/0.988} \\ 
		Rain200H  &     {28.16/0.920}    &27.63/0.872 &27.03/0.955& - &   \textcolor{blue}{29.31/0.901} & \textcolor{red}{30.70/0.922} \\ 
		Rain800    &   26.32/0.902       &25.93/0.832  &25.64/0.833 & -& \textcolor{blue}{28.39/0.876} & \textcolor{red}{28.42/0.876}\\
		Rain1200  &    -   &{32.91/0.916} &20.12/0.691& - & \textcolor{blue}{33.17/0.922} & \textcolor{red}{33.70/0.928}\\ 
% 		Rain1400 \cite{fu2017removing} & 27.55/0.853                & 28.57/0.891                     & 28.23/0.924                    & \textcolor{red}{32.60/0.946}                & 31.33/0.918                     & \textcolor{blue}{31.95/0.926}\\
		SPA-Data  &   -    &\textcolor{blue}{44.89/0.989}    & 17.75/0.515 &-& {42.81/0.986} & \textcolor{red}{46.88/0.991}\\ 
		Params &     5,230,214    &3,637,303  &115,333,723& 115,333,723  & 2,158,586 & 129,539,018 \\  % & 129,539,018
% 		Running Time & 0.00715               & 0.0172                     & 0.0077                    & 0.0290                        & 0.0009         & 0.09675       & -        & 0.1519 & 0.0893 \\
	
	\bottomrule
	
	\end{tabular*} }
	}
	\smallskip
	\label{table:syn}
\end{table*}

\begin{table}[h]
	\tabcolsep 0.05in{\scriptsize{}}
	\caption{Synthetic and real-world datasets}
	\vspace{0.15cm}
    \begin{tabular*}{\tblwidth}{@{}LLLL@{}}		\toprule
		Datasets                         & Train           & Test           & Type \\ \midrule
		Rain200L \cite{yang2017deep}      & 1,800                  & 200                   & synthetic\\
		Rain200H \cite{yang2017deep}      & 1,800                  & 200                   & synthetic\\
		Rain800 \cite{zhang2019image}     & 700                    & 100                   & synthetic\\
		Rain1200 \cite{zhang2018density}  & 12,000                 & 1,200                 & synthetic\\
		RainCityscapes \cite{Cordts2016Cityscapes,hu2019depth}  & 9,432         & 1,188                  & synthetic\\
	
		SPA-Data \cite{wang2019spatial}      & 640k                    & 1,000                 & real-world\\
		Yang \textit{et al.} \cite{yang2017deep}   & -                      & 15                    & real-world\\
		DQA \cite{wu2020subjective}    & -                      & 206                    & real-world \\ 

		 Raindrop \cite{qian2018attentive}   & 861                      & 58 (A)/249 (B)             & real-world \\  \bottomrule

	\end{tabular*}
	\smallskip
	\label{tab:dataset}
\end{table}

\section{Experiments}

In this section, we present the dataset used in this study and describe the details of the experimental setting. We present two versions of the proposed MCW-Net: a small model and a large model. The architecture of the two models is same except for the number of channels. The small model has eight times fewer channels than the large model.
We conduct a quantitative and qualitative evaluation of the proposed method and compare its performance with state-of-the-art methods. An ablation study is conducted to confirm the significance of each component introduced in Section \ref{sec:proposed}.

\label{sec:exp}

\subsection{Datasets and Evaluation Metrics}

Five synthetic datasets (Rain200H, Rain200L, Rain800, Rain1200, RainCityscapes) and three real-world datasets (SPA-Data, Yang \textit{et al.} \cite{yang2017deep}, DQA \cite{wu2020subjective})  are used to evaluate the performance of the proposed method. As pointed out by Ren \textit{et al.} \cite{ren2019progressive}, certain overlaps of background exist between the training and test datasets in the Rain100H and Rain100L datasets. Therefore, we evaluate our model using the updated Rain200H and Rain200L datasets, which do not share backgrounds with the corresponding training datasets.
Because the absence of ground truth data makes quantitative evaluation impossible, the real-world dataset of \cite{yang2017deep} is evaluated qualitatively using the Rain200H-trained weights. 
In addition, Raindrop \cite{qian2018attentive} dataset is used to evaluate the raindrop removal performance of the proposed method. We compare the performance of the proposed method with nine state-of-the-art single-image deraining methods. \\
We employ PSNR and SSIM \cite{wang2004image} metrics for quantitative quality
assessment. All the PSNRs reported in the following experimental results are calculated for RGB channels. Some previous works filter the derained RGB images into YCbCr space and then evaluate PSNR only for the Y channel to focus on the luminance. However, because most of the high-level vision algorithms commonly receive the RGB image as an input, we consider evaluation of well-recovered rainy image in RGB space is more appropriate and helpful for other vision tasks. Additionally, we employ the dedicated B-FEN metric \cite{wu2020subjective} to measure the deraining quality of deraining algorithms. 
% For more information on datasets and traininig details, please refer the supplementary material. 

\subsection{Datasets and Experiment Details}

For all the datasets, we randomly crop $256\times256$ patch from each input image. During the training, we set the batch size to 4 and use the Adam optimizer.
For the large model, we set the learning rate to be $10^{-4}$ and train our model for 200 epochs on the Rain200H, Rain200L, and Rain800 datasets, 100 epochs on the Rain1200 and the RainCityscapes datasets, 3 epochs on the SPA-Data dataset, and 500 epochs on the Raindrop dataset. For the small model, we set the learning rate to be $5\times10^{-4}$  and train our model for 500 epochs on the Rain200H, Rain200L, and Rain800 datasets, 100 epochs on the Rain1200 and the RainCityscapes datasets, 5 epochs on the SPA-Data dataset, and 750 epochs on the Raindrop dataset.

\begin{figure*}
	\centering
	\setlength{\tabcolsep}{0pt}
	\footnotesize{
	\begin{tabular}{cccccccclcccccccclcccccccclcccccccclcccccccclccccccccl}
		\multicolumn{3}{c}{\includegraphics[width=0.16\textwidth]{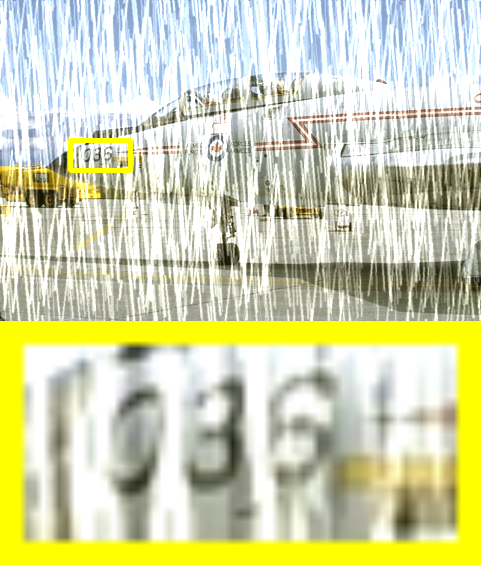}}\ &
		\multicolumn{3}{c}{\includegraphics[width=0.16\textwidth]{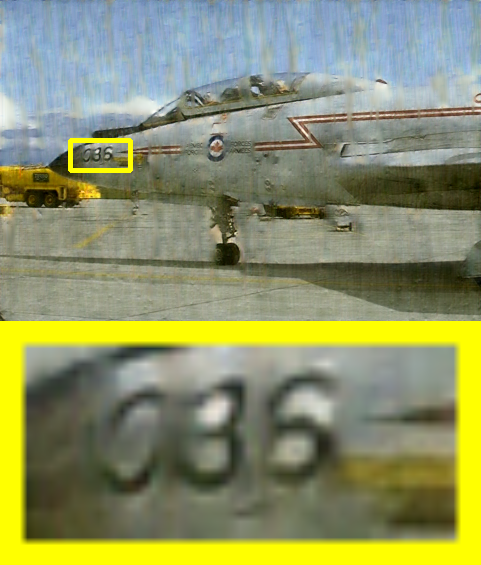}}\ &
		\multicolumn{3}{c}{\includegraphics[width=0.16\textwidth]{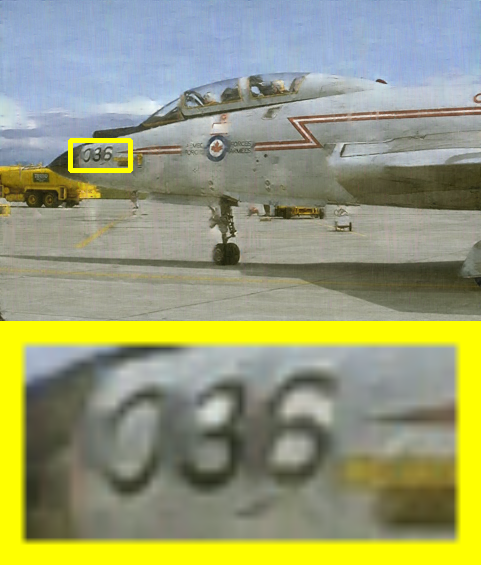}}\ &
		\multicolumn{3}{c}{\includegraphics[width=0.16\textwidth]{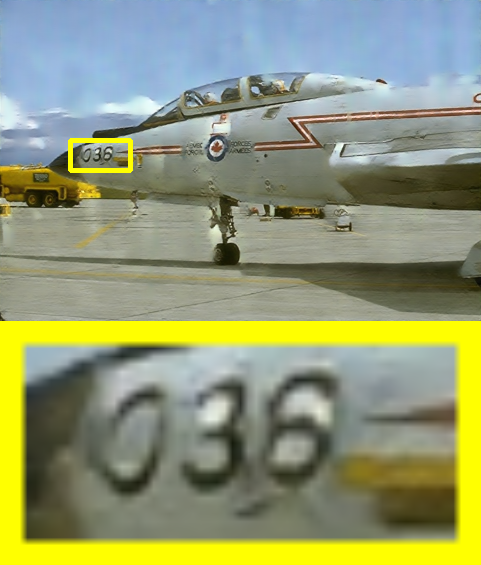}}\ &
		\multicolumn{3}{c}{\includegraphics[width=0.16\textwidth]{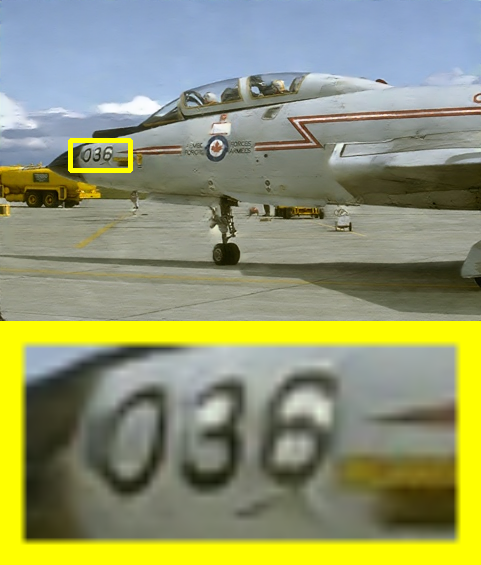}}\ &
		\multicolumn{3}{c}{\includegraphics[width=0.16\textwidth]{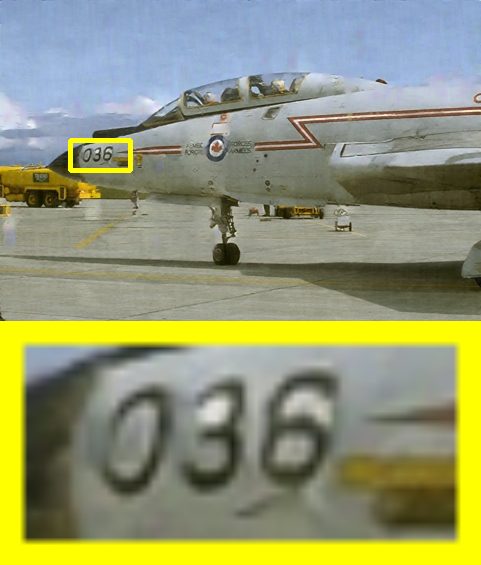}}\\
		
		\multicolumn{3}{c}{(a) Rainy image} &
		\multicolumn{3}{c}{(b) JORDER~\cite{yang2017deep}} &
		\multicolumn{3}{c}{(c) RESCAN~\cite{li2018recurrent}} &
		\multicolumn{3}{c}{(d) SPANet~\cite{wang2019spatial}} &
		\multicolumn{3}{c}{(e) PReNet~\cite{ren2019progressive} } &
		\multicolumn{3}{c}{(f) ReHEN~\cite{yang2019single}} \\
		\multicolumn{3}{c}{\includegraphics[width=0.16\textwidth]{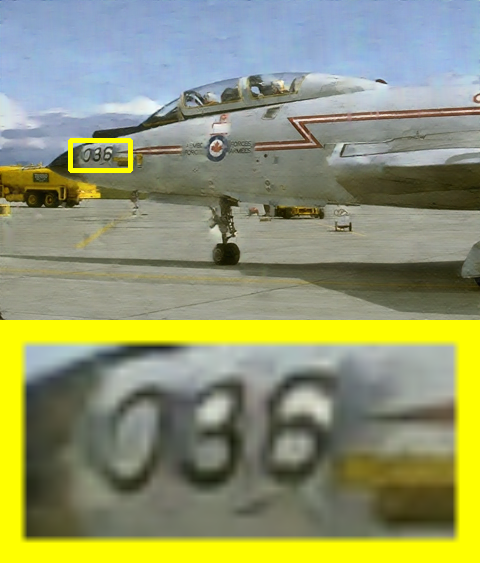}}\ &
		\multicolumn{3}{c}{\includegraphics[width=0.16\textwidth]{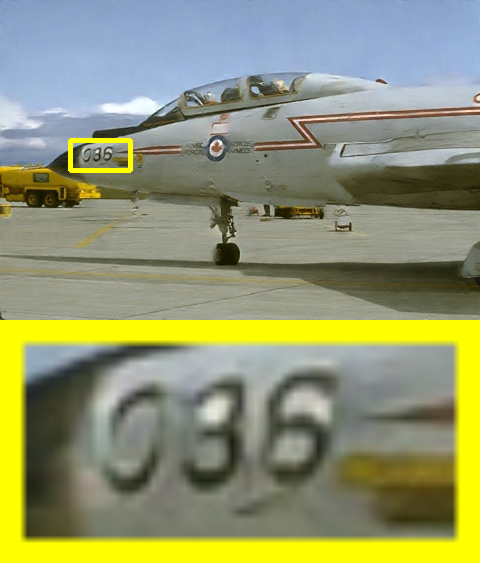}}\ &
		\multicolumn{3}{c}{\includegraphics[width=0.16\textwidth]{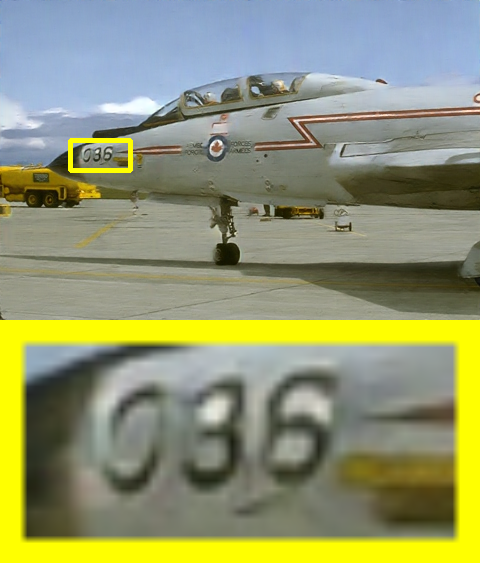}}\ &
		\multicolumn{3}{c}{\includegraphics[width=0.16\textwidth]{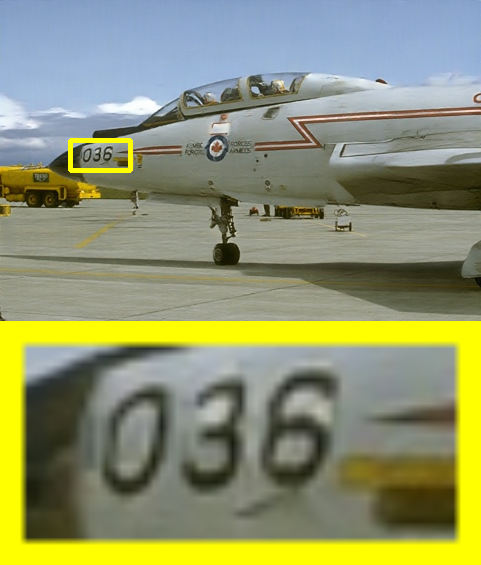}}\ &
		\multicolumn{3}{c}{\includegraphics[width=0.16\textwidth]{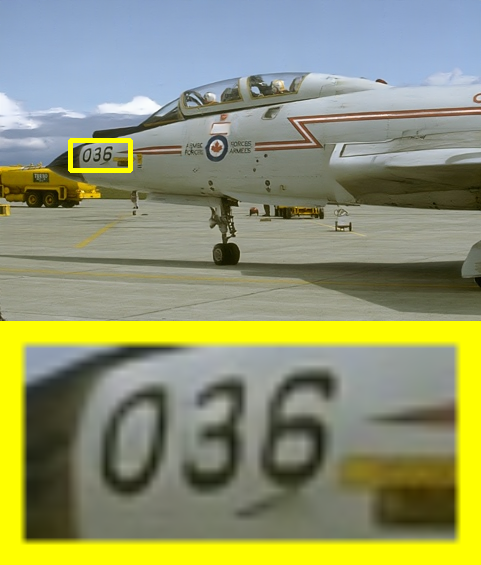}}\ &
		\multicolumn{3}{c}{\includegraphics[width=0.16\textwidth]{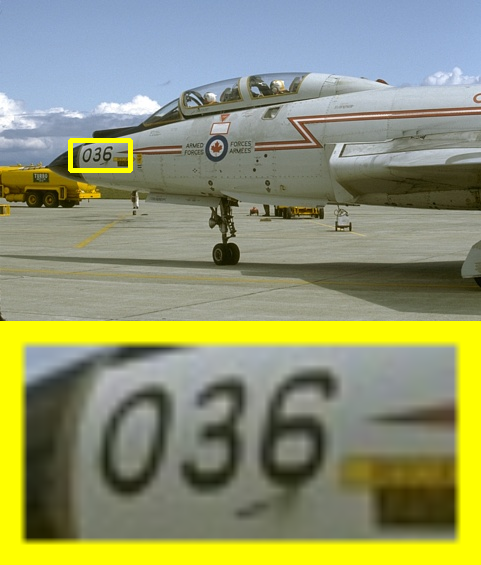}}\\
		\multicolumn{3}{c}{(g) RCDNet~\cite{wang2020model}} &
		\multicolumn{3}{c}{(h) MPRNet~\cite{zamir2021multi}} &
		\multicolumn{3}{c}{(h) IPT \cite{chen2021pre} } &
		\multicolumn{3}{c}{(i) MCW-Net(small)} &
		\multicolumn{3}{c}{(j) MCW-Net(large)} &
		\multicolumn{3}{c}{(k) GT} \\
		
		\multicolumn{3}{c}{\includegraphics[width=0.16\textwidth]{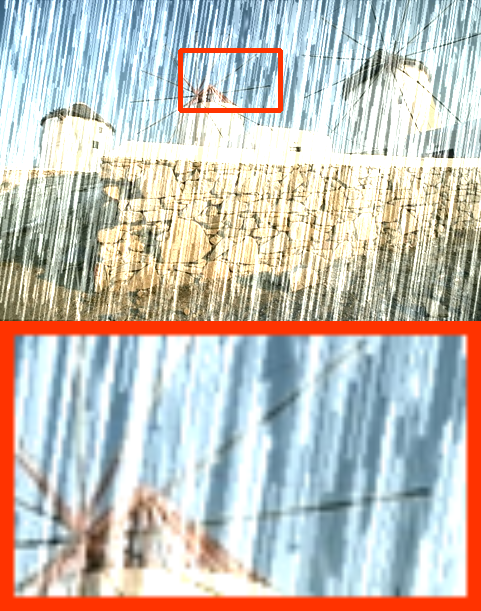}}\ &
		\multicolumn{3}{c}{\includegraphics[width=0.16\textwidth]{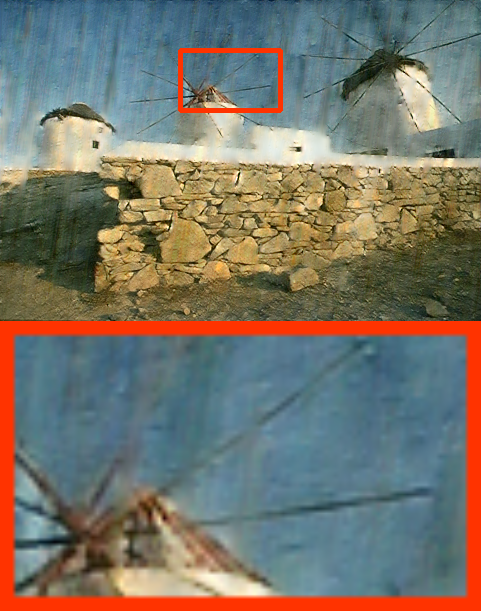}}\ &
		\multicolumn{3}{c}{\includegraphics[width=0.16\textwidth]{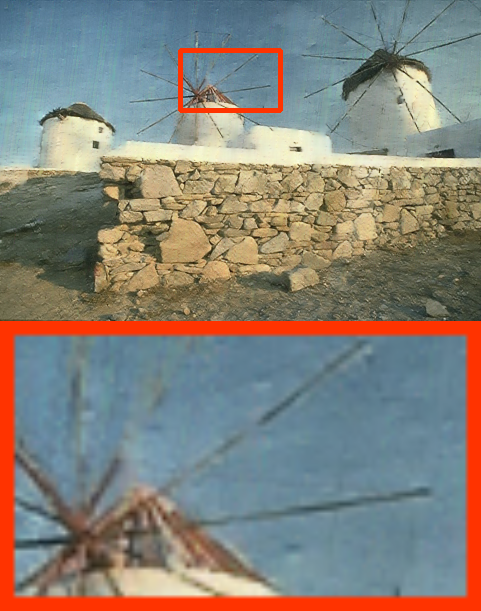}}\ &
		\multicolumn{3}{c}{\includegraphics[width=0.16\textwidth]{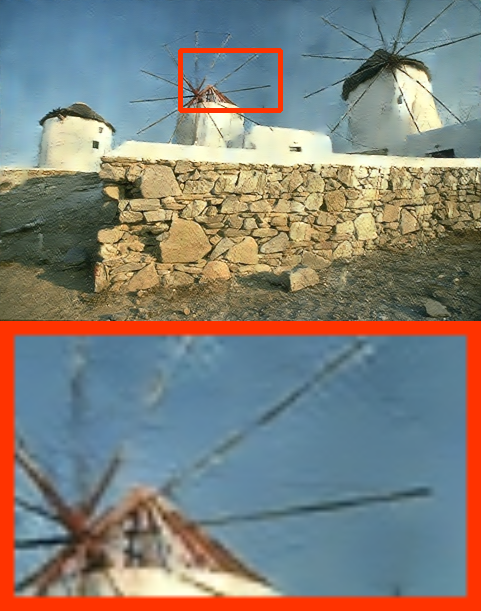}}\ &
		\multicolumn{3}{c}{\includegraphics[width=0.16\textwidth]{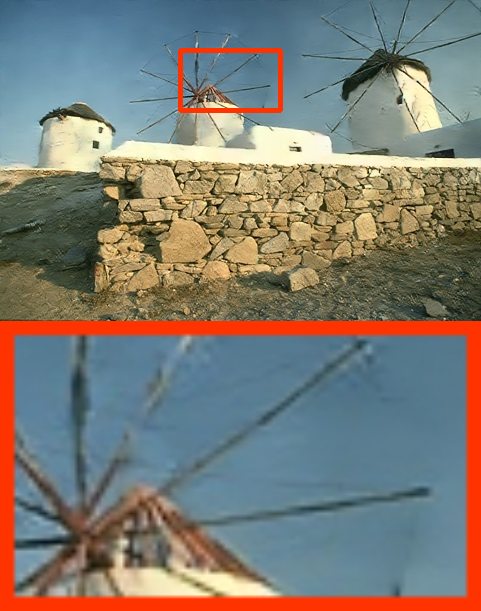}}\ &
		\multicolumn{3}{c}{\includegraphics[width=0.16\textwidth]{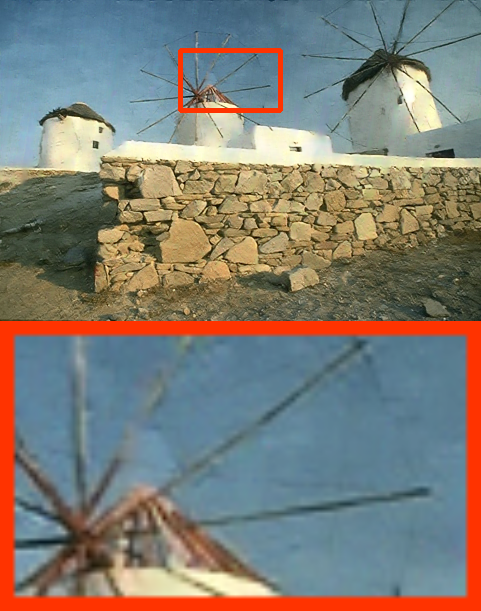}}\\
			
		\multicolumn{3}{c}{(a) Rainy image} &
		\multicolumn{3}{c}{(b) JORDER~\cite{yang2017deep}} &
		\multicolumn{3}{c}{(c) RESCAN~\cite{li2018recurrent}} &
		\multicolumn{3}{c}{(d) SPANet~\cite{wang2019spatial}} &
		\multicolumn{3}{c}{(e) PReNet~\cite{ren2019progressive} } &
		\multicolumn{3}{c}{(f) ReHEN~\cite{yang2019single}} \\
		\multicolumn{3}{c}{\includegraphics[width=0.16\textwidth]{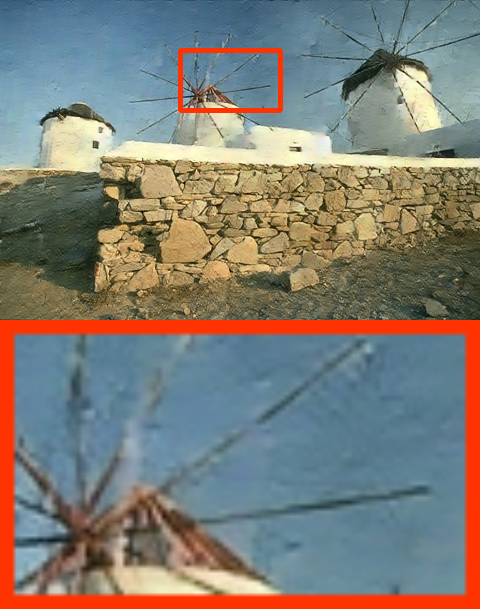}}\ &
		\multicolumn{3}{c}{\includegraphics[width=0.16\textwidth]{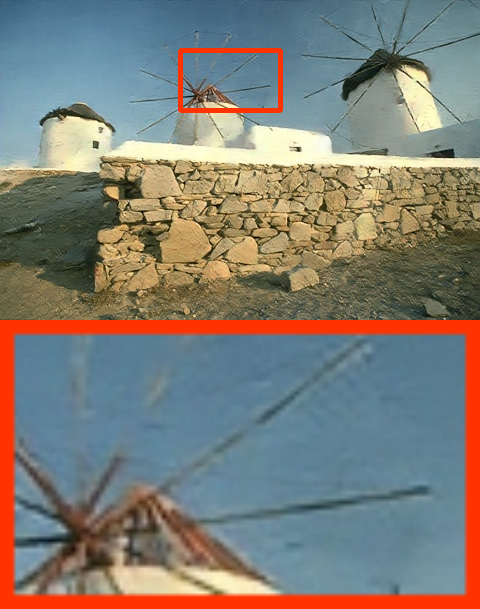}}\ &
		\multicolumn{3}{c}{\includegraphics[width=0.16\textwidth]{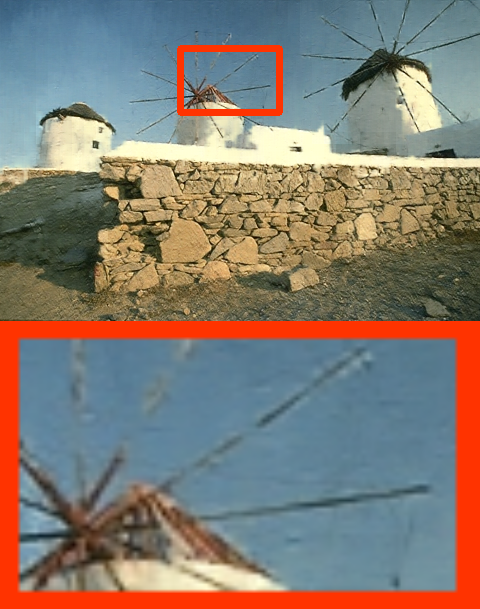}}\ &
		\multicolumn{3}{c}{\includegraphics[width=0.16\textwidth]{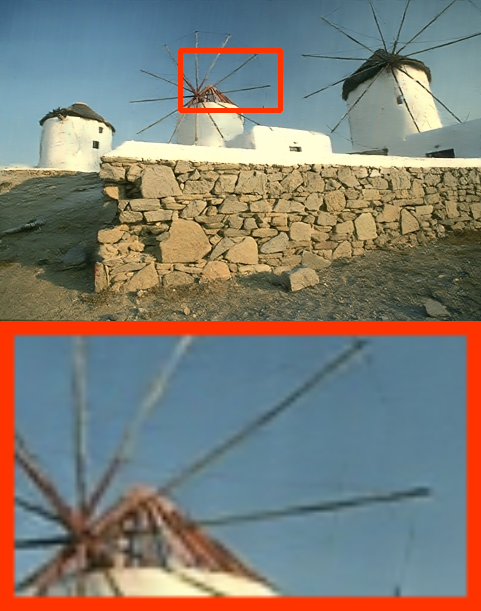}}\ &
		\multicolumn{3}{c}{\includegraphics[width=0.16\textwidth]{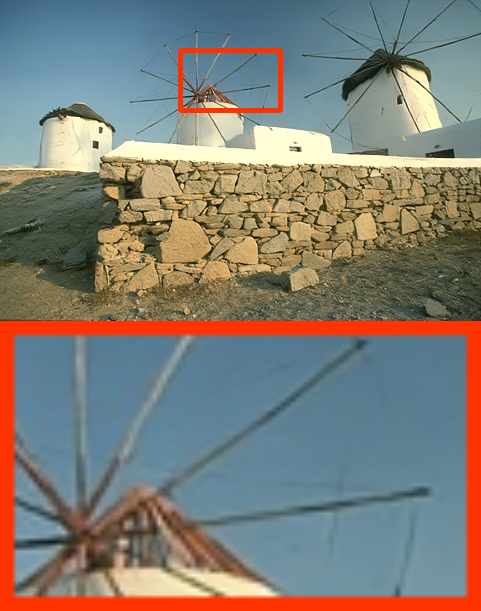}}\ &
		\multicolumn{3}{c}{\includegraphics[width=0.16\textwidth]{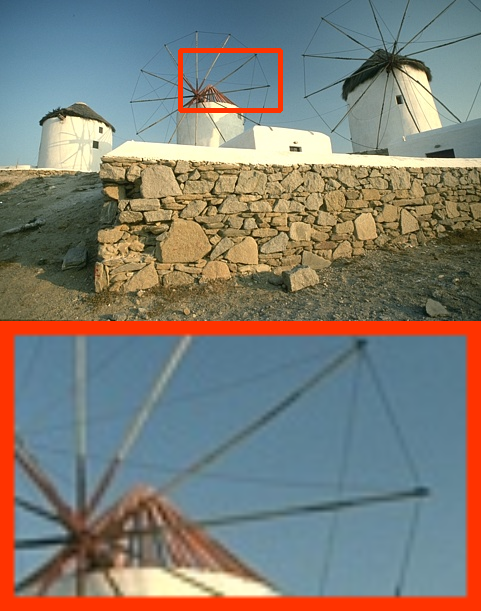}}\\
		
		\multicolumn{3}{c}{(g) RCDNet~\cite{wang2020model}} &
		\multicolumn{3}{c}{(h) MPRNet~\cite{zamir2021multi}} &
		\multicolumn{3}{c}{(h) IPT \cite{chen2021pre} } &
		\multicolumn{3}{c}{(i) MCW-Net(small)} &
		\multicolumn{3}{c}{(j) MCW-Net(large)} &
		\multicolumn{3}{c}{(k) GT} \\

	\end{tabular}}
	\caption{Results obtained via several state-of-the-art methods on the Rain200H \cite{yang2017deep} images. The outputs of MCW-Net exhibit no traces of rain streaks on both image samples. MCW-Net also recovers the most detailed images.}
	\label{fig:results rain200h}
\end{figure*}

\begin{figure*}
	\centering
	\setlength{\tabcolsep}{0pt}
	\footnotesize{
	\begin{tabular}{cccccccclcccccccclcccccccclcccccccclccccccccl}
		\multicolumn{3}{c}{\includegraphics[width=0.19\textwidth]{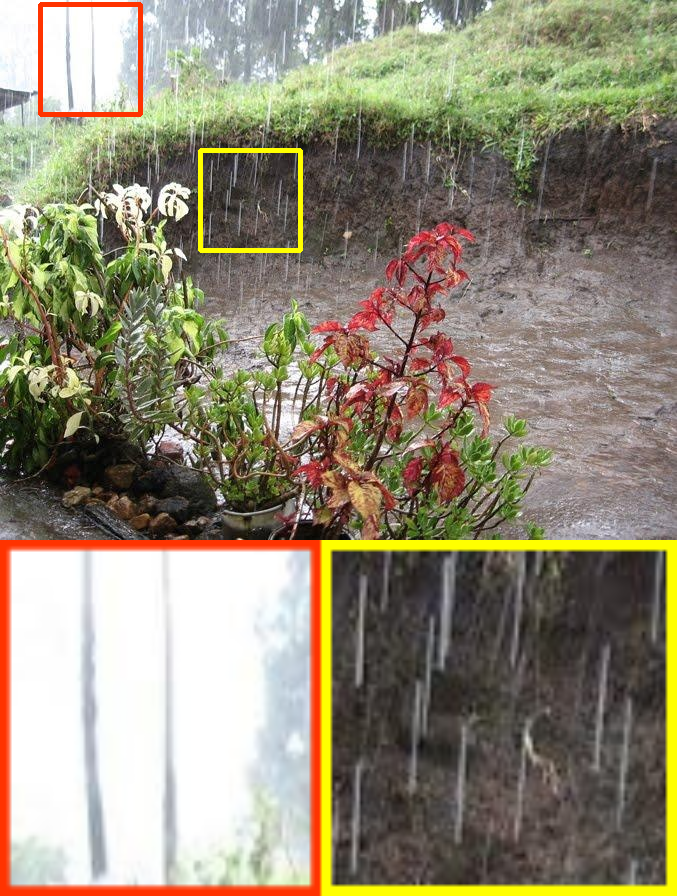}}\ &
		\multicolumn{3}{c}{\includegraphics[width=0.19\textwidth]{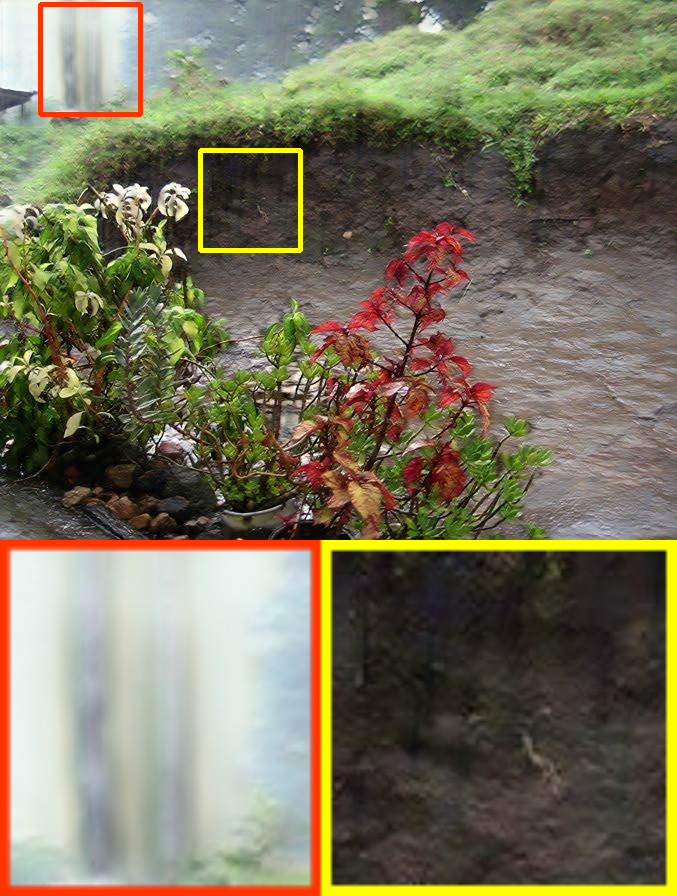}}\ &
		\multicolumn{3}{c}{\includegraphics[width=0.19\textwidth]{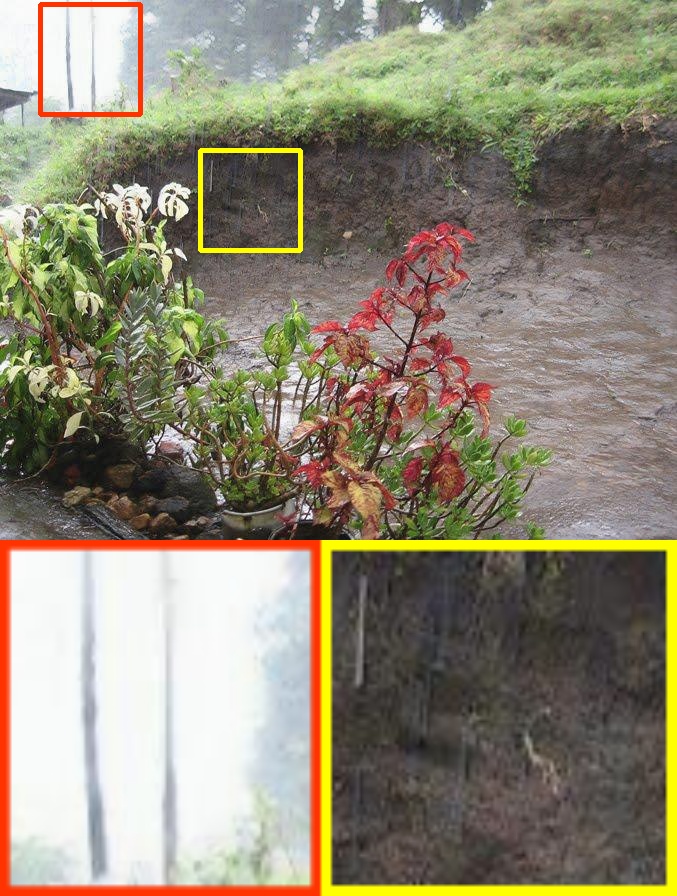}}\ &
		\multicolumn{3}{c}{\includegraphics[width=0.19\textwidth]{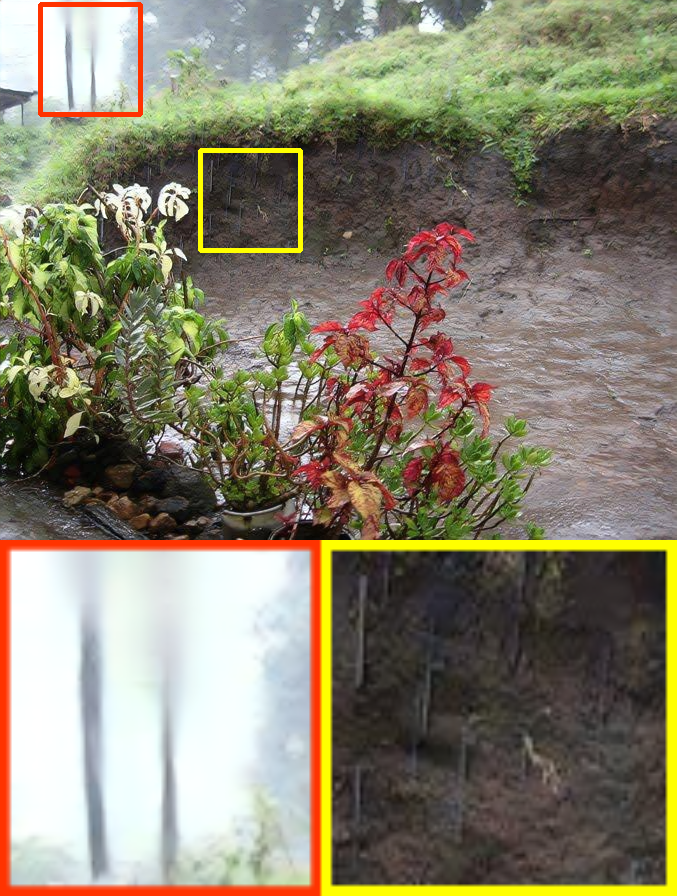}}\ &
		\multicolumn{3}{c}{\includegraphics[width=0.19\textwidth]{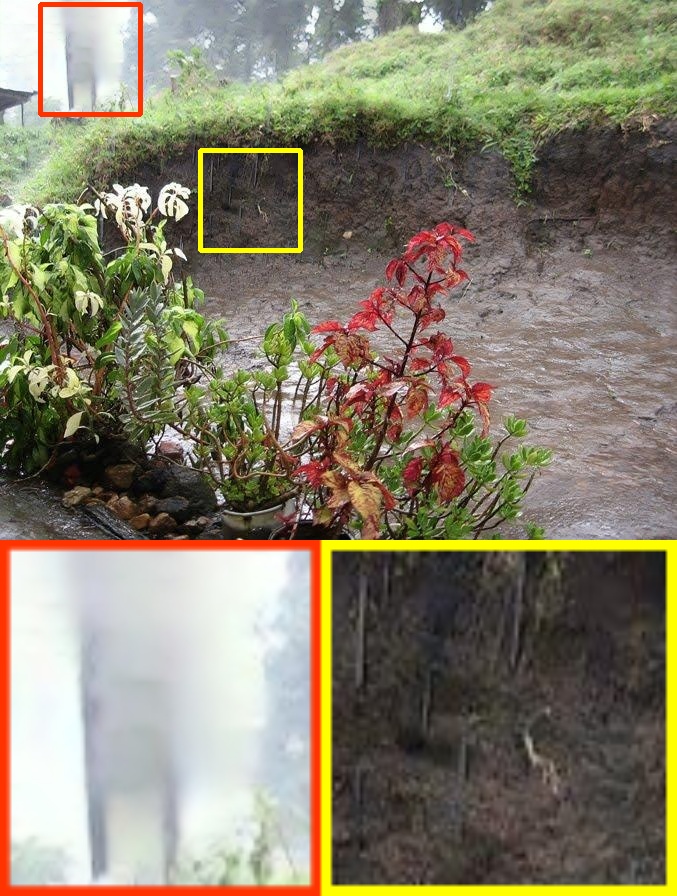}}\\

		\multicolumn{3}{c}{(a) Rainy image} &
		\multicolumn{3}{c}{(b) JORDER~\cite{yang2017deep}} &
		\multicolumn{3}{c}{(c) RESCAN~\cite{li2018recurrent}} &
		\multicolumn{3}{c}{(d) SPANet~\cite{wang2019spatial}} &
		\multicolumn{3}{c}{(e) PReNet~\cite{ren2019progressive} } \\

		\multicolumn{3}{c}{\includegraphics[width=0.19\textwidth]{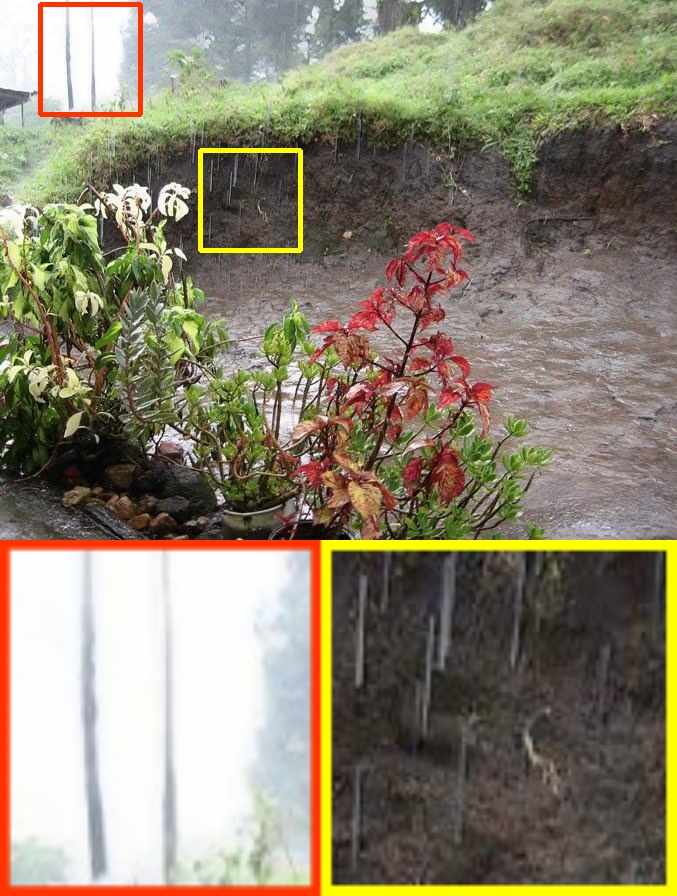}}\ &
		\multicolumn{3}{c}{\includegraphics[width=0.19\textwidth]{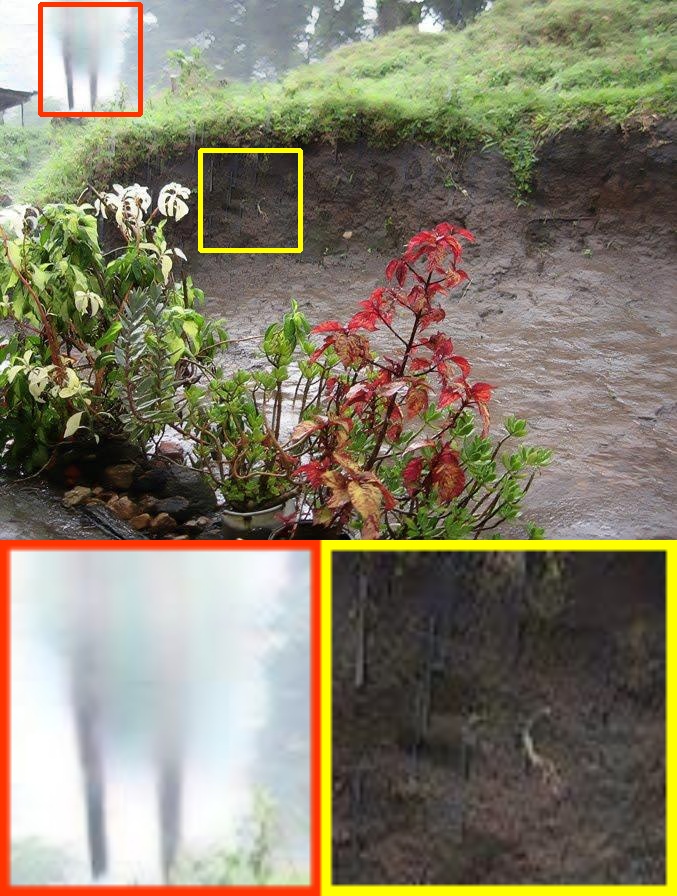}}\ &
		\multicolumn{3}{c}{\includegraphics[width=0.19\textwidth]{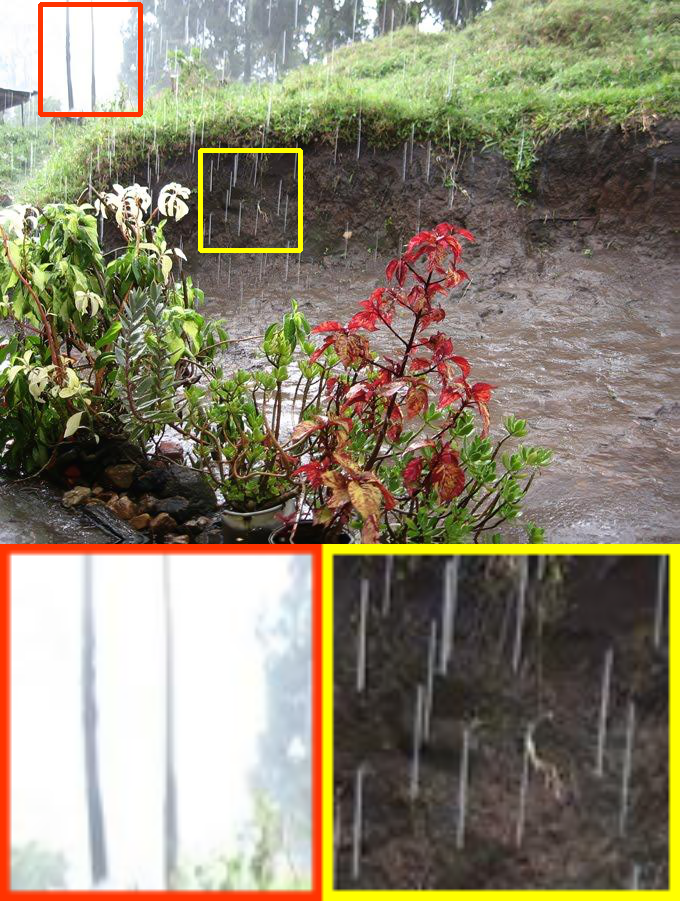}}\ &
		\multicolumn{3}{c}{\includegraphics[width=0.19\textwidth]{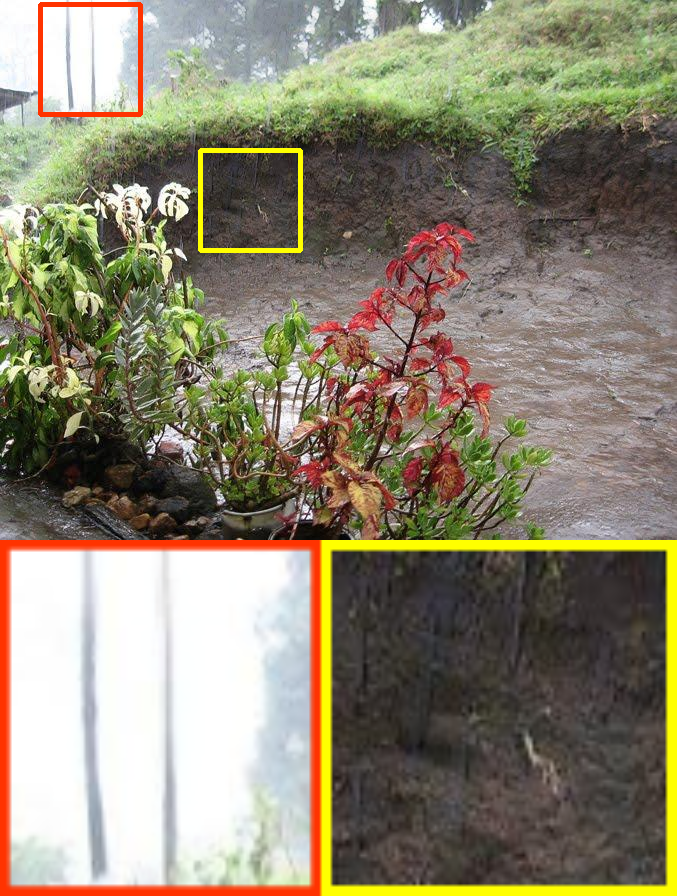}}\ &
		\multicolumn{3}{c}{\includegraphics[width=0.19\textwidth]{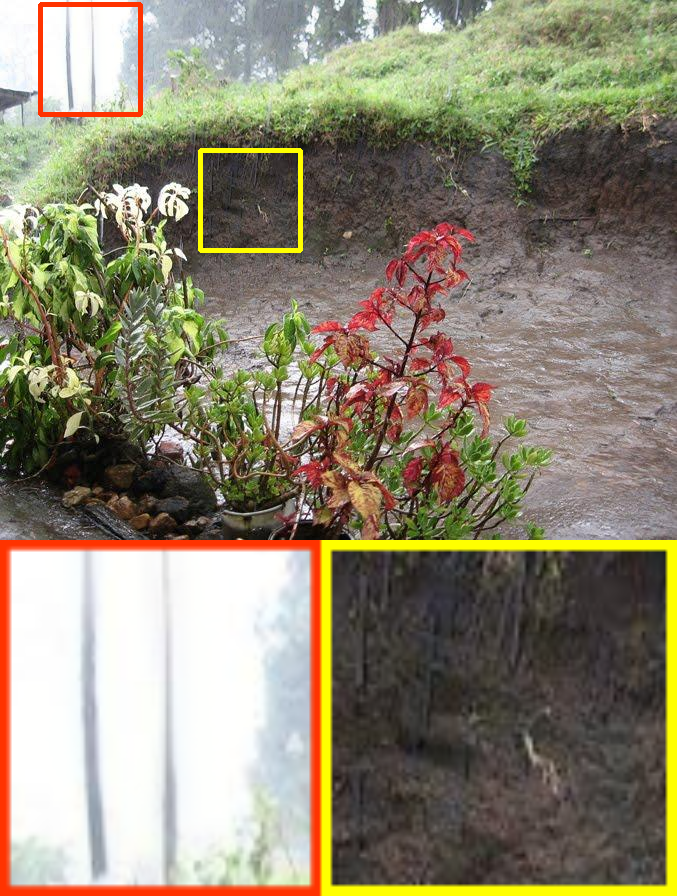}}\\

		\multicolumn{3}{c}{(f) ReHEN~\cite{yang2019single}} &		
		\multicolumn{3}{c}{(g) RCDNet~\cite{wang2020model}} &
		\multicolumn{3}{c}{(h) MPRNet~\cite{zamir2021multi}} &
		\multicolumn{3}{c}{(i) MCW-Net (small)} &
		\multicolumn{3}{c}{(j) MCW-Net (large)} \\
		
		\multicolumn{3}{c}{\includegraphics[width=0.19\textwidth]{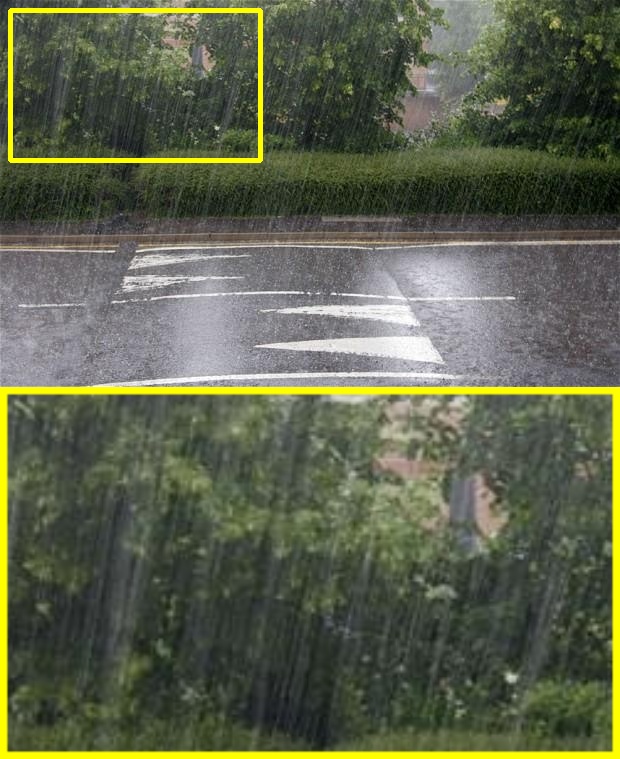}}\ &
		\multicolumn{3}{c}{\includegraphics[width=0.19\textwidth]{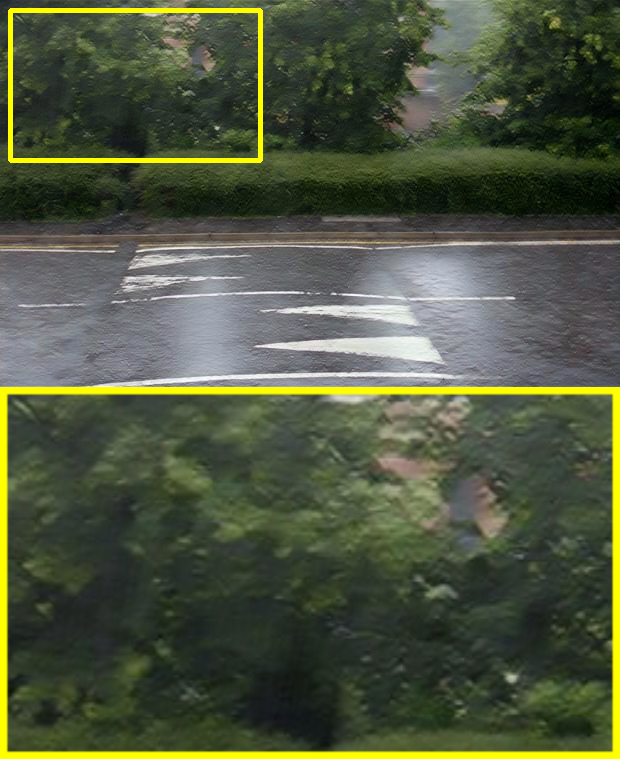}}\ &
		\multicolumn{3}{c}{\includegraphics[width=0.19\textwidth]{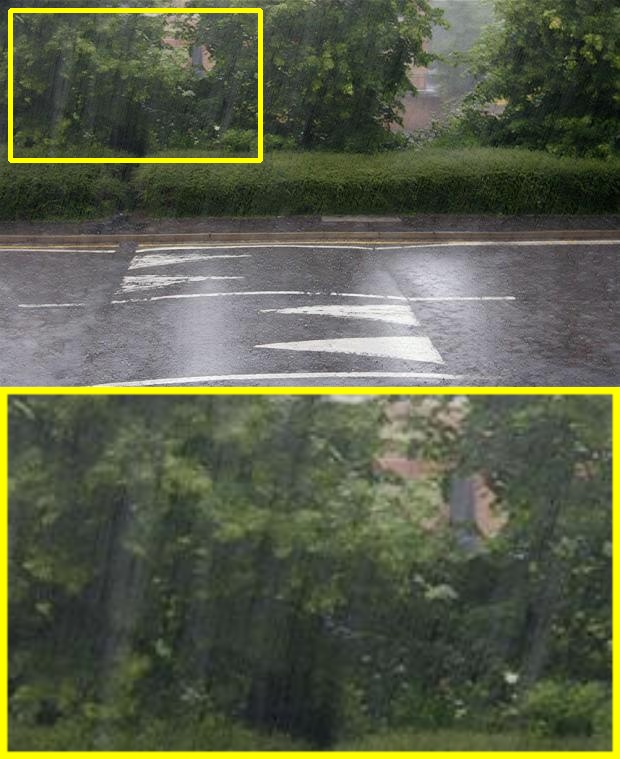}}\ &
		\multicolumn{3}{c}{\includegraphics[width=0.19\textwidth]{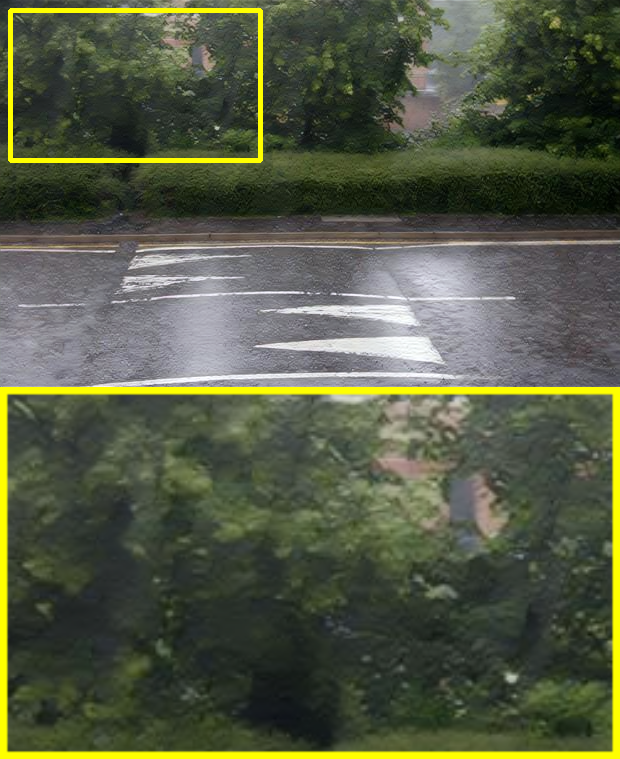}}\ &
		\multicolumn{3}{c}{\includegraphics[width=0.19\textwidth]{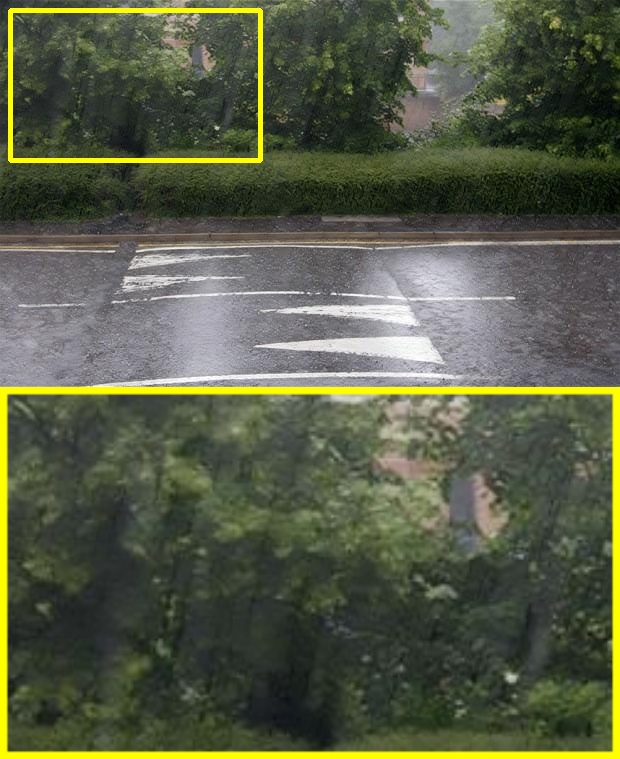}}\\

		\multicolumn{3}{c}{(a) Rainy image} &
		\multicolumn{3}{c}{(b) JORDER~\cite{yang2017deep}} &
		\multicolumn{3}{c}{(c) RESCAN~\cite{li2018recurrent}} &
		\multicolumn{3}{c}{(d) SPANet~\cite{wang2019spatial}} &
		\multicolumn{3}{c}{(e) PReNet~\cite{ren2019progressive} } \\

		\multicolumn{3}{c}{\includegraphics[width=0.19\textwidth]{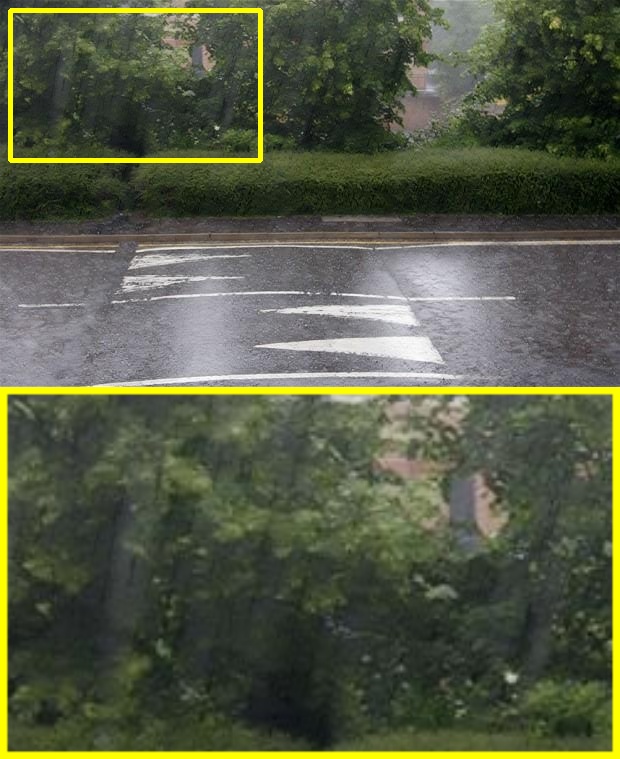}}\ &
		\multicolumn{3}{c}{\includegraphics[width=0.19\textwidth]{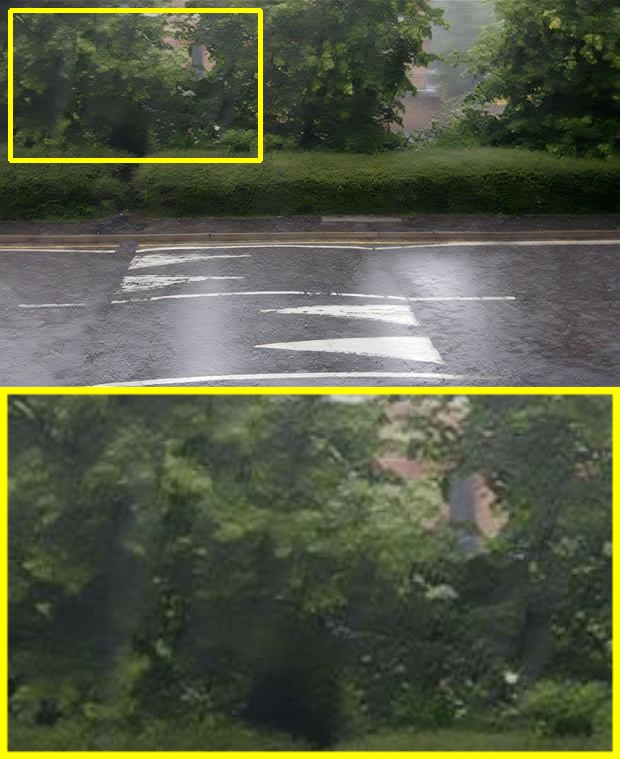}}\ &
		\multicolumn{3}{c}{\includegraphics[width=0.19\textwidth]{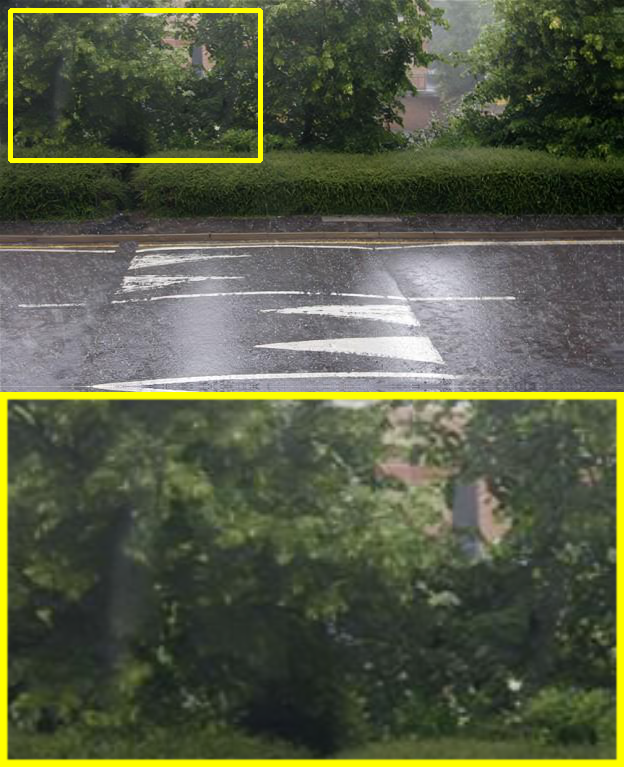}}\ &
		\multicolumn{3}{c}{\includegraphics[width=0.19\textwidth]{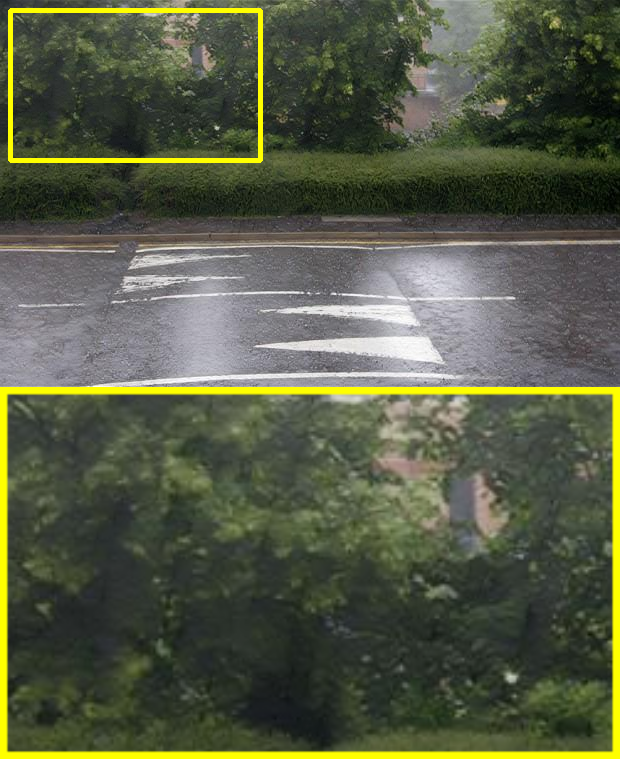}}\ &
		\multicolumn{3}{c}{\includegraphics[width=0.19\textwidth]{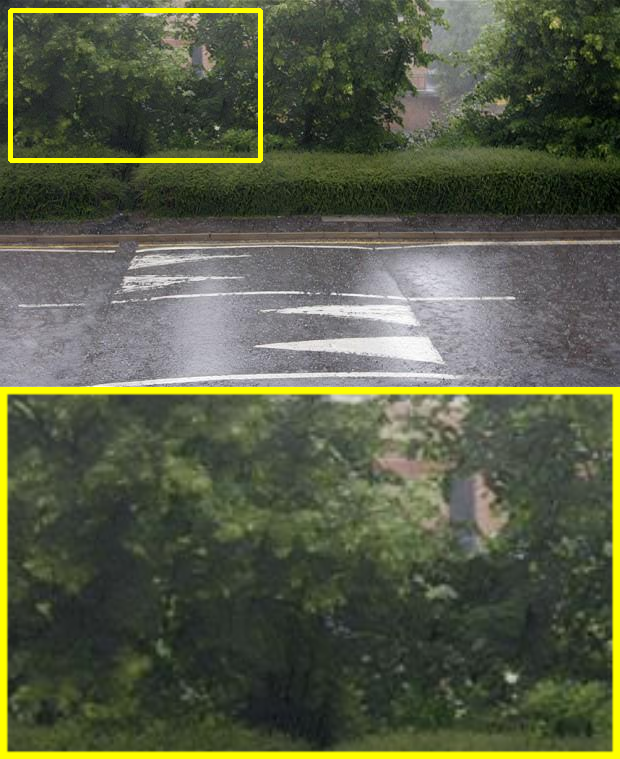}}\\

		\multicolumn{3}{c}{(f) ReHEN~\cite{yang2019single}} &		
		\multicolumn{3}{c}{(g) RCDNet~\cite{wang2020model}} &
		\multicolumn{3}{c}{(h) MPRNet~\cite{zamir2021multi}} &
		\multicolumn{3}{c}{(i) MCW-Net (small)} &
		\multicolumn{3}{c}{(j) MCW-Net (large)} \\
	\end{tabular}}
	\caption{Results obtained via several state-of-the-art methods on the Yang \textit{et al.} \cite{yang2017deep} images. Among state-of-the-art methods, MCW-Net is the only one that restore the detail of the images while removing the rain streaks.}
	\label{fig:results real}
\end{figure*}

\begin{figure*}
	\centering
	\setlength{\tabcolsep}{0pt}
	\footnotesize{
	\begin{tabular}{cccccccclcccccccclcccccccclccccccccl}

		\multicolumn{3}{c}{\includegraphics[width=0.24\textwidth]{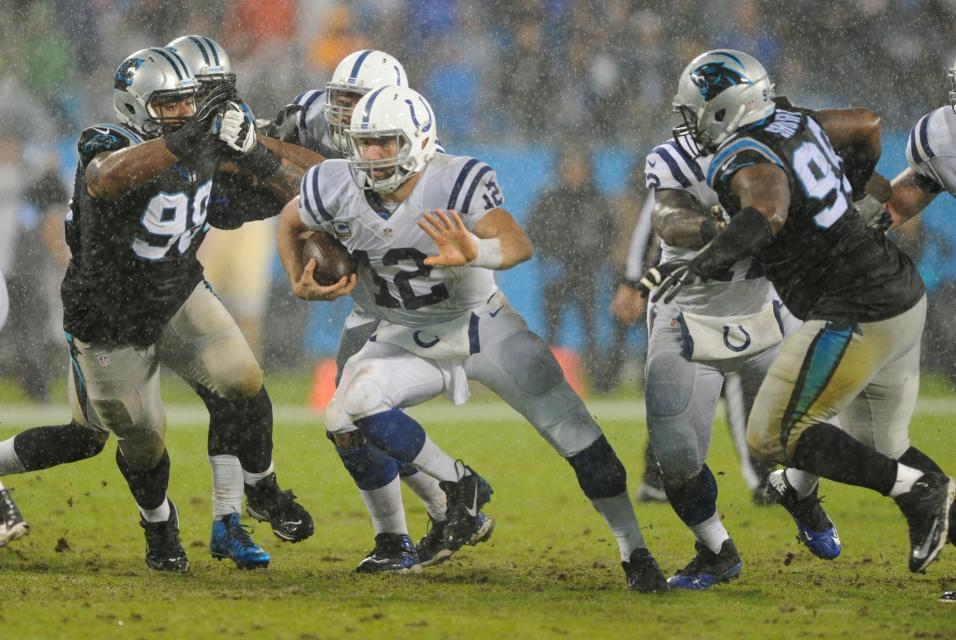}}\ &
		\multicolumn{3}{c}{\includegraphics[width=0.24\textwidth]{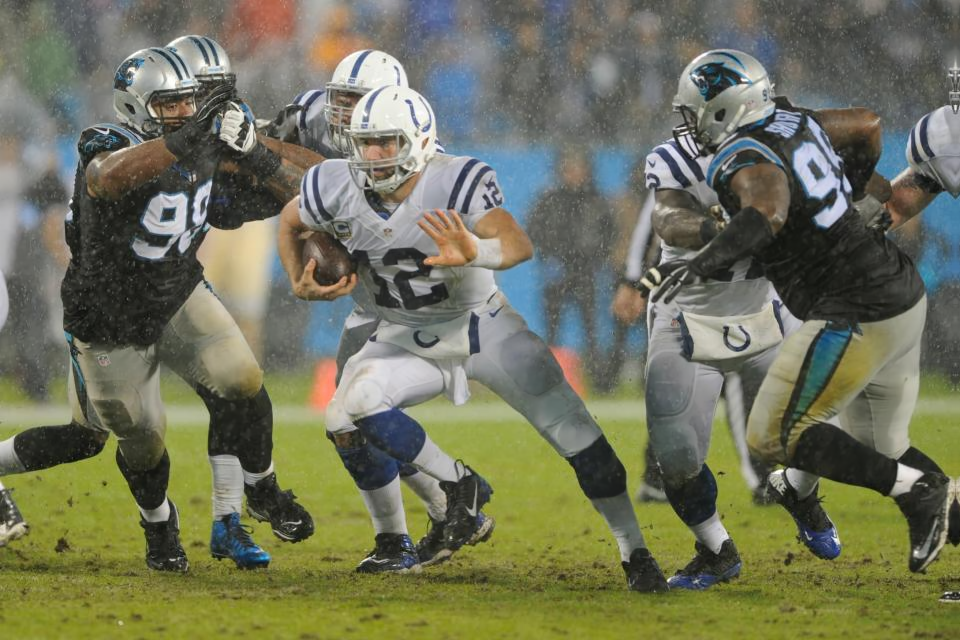}}\ &
		\multicolumn{3}{c}{\includegraphics[width=0.24\textwidth]{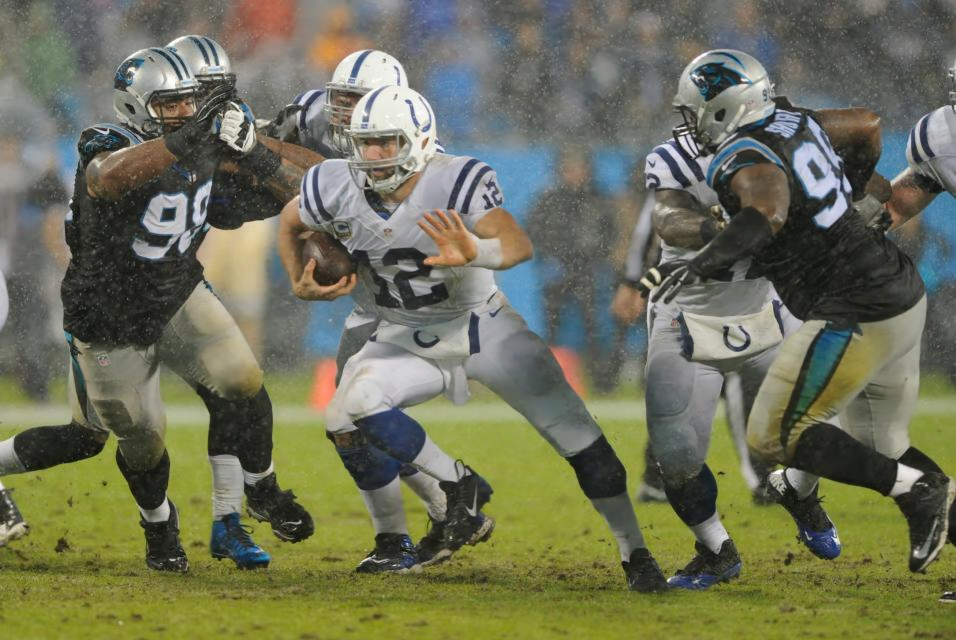}}\ &
		\multicolumn{3}{c}{\includegraphics[width=0.24\textwidth]{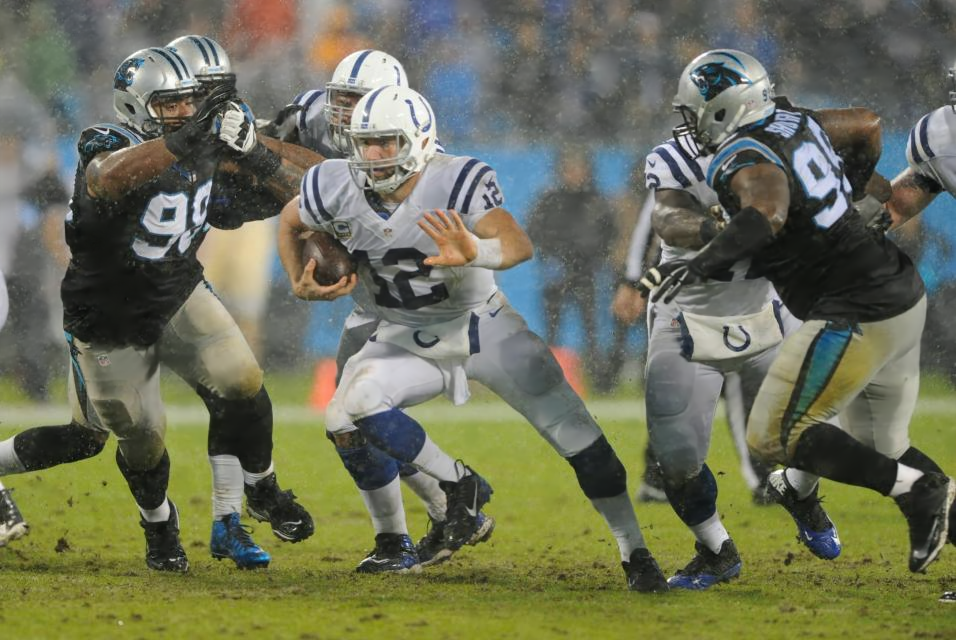}}\\ &
		\\
		
		\multicolumn{3}{c}{(a) Rainy image} &
		\multicolumn{3}{c}{(b) MPRNet~\cite{zamir2021multi}} &
		\multicolumn{3}{c}{(c) RCDNet~\cite{wang2020model}} &
		\multicolumn{3}{c}{(d) PReNet~\cite{ren2019progressive}} & \\

		\multicolumn{3}{c}{} & 
		\multicolumn{3}{c}{\includegraphics[width=0.24\textwidth]{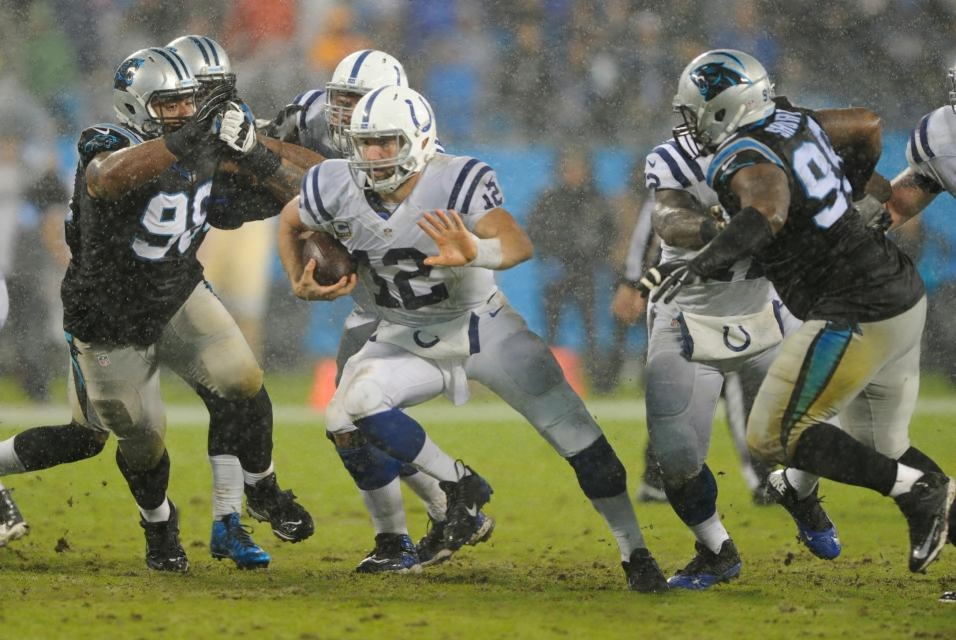}}\ &
		\multicolumn{3}{c}{\includegraphics[width=0.24\textwidth]{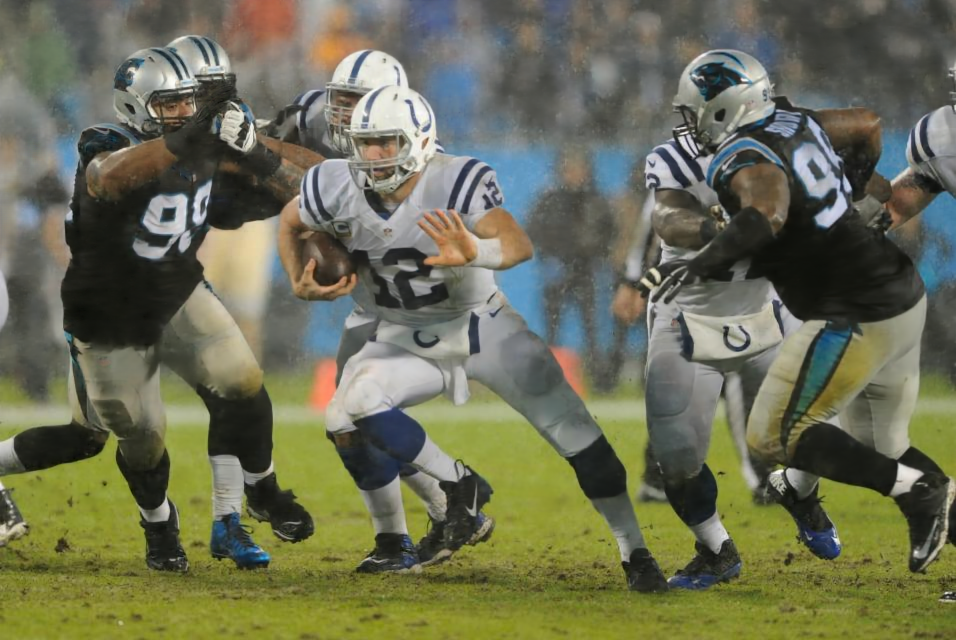}}\ &
		\multicolumn{3}{c}{\includegraphics[width=0.24\textwidth]{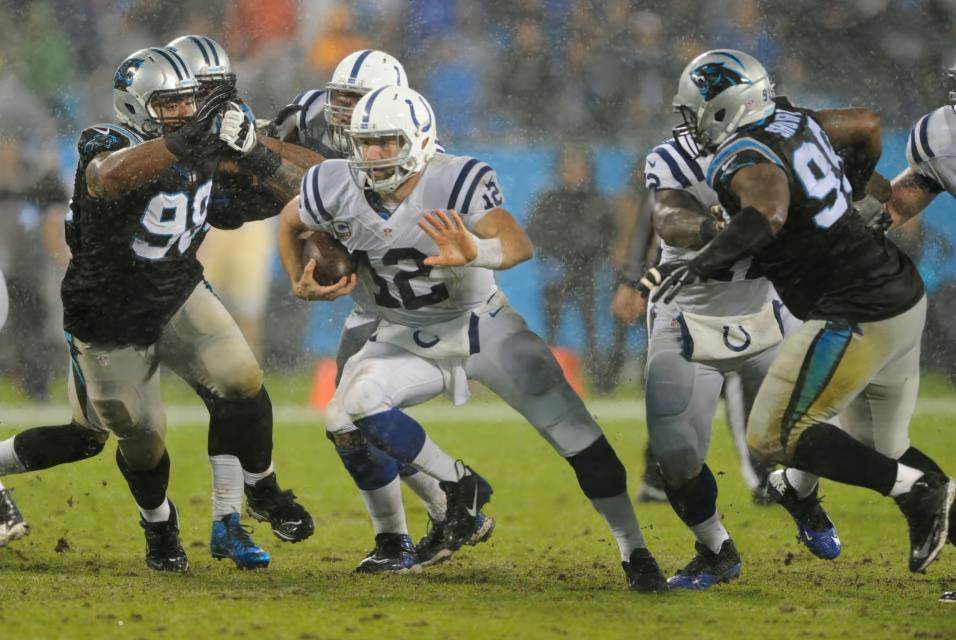}}\\ &
        \\ 
        \multicolumn{3}{c}{} & 
		\multicolumn{3}{c}{(e) SPANet~\cite{wang2019spatial} } &
		\multicolumn{3}{c}{(g) MCW-Net (small)} &
		\multicolumn{3}{c}{(j) MCW-Net (large)} & \\
	\end{tabular}}
	\caption{Results obtained via several state-of-the-art methods on the DQA images. Among state-of-the-art methods, MCW-Net is the only one that restore the detail of the images while removing the rain streaks. Based on the area of left-most person in (g) and (j), the small version removed rain better than the larger version, consistent with the results of the B-FEN score. However, the small version remove all the wrinkles on clothes, and the large version preserve them, so the large version achieves a better restoration of details.}
	\label{fig:results dqa}
\end{figure*}

\subsection{Evaluations}
\label{subsec:Evaluations}

\subsubsection{Results on Synthetic Datasets}
\label{subsec:results_synthetic}

% \textcolor{blue}{
% As mentioned in Section \ref{sec:exp}, the proposed MCW-Net is evaluated on four synthetic datasets \cite{ wang2019spatial, yang2017deep, zhang2019image}, and the performance is compared to eight state-of-the art methods \cite{deng2020detail,li2018recurrent, ren2019progressive,wang2020model, wang2019spatial, yang2017deep, yang2019single,zamir2021multi}. The quantitative results on the synthetic datasets are presented in Table \ref{table:syn}. As is evident from the data, the proposed MCW-Net achieves remarkable the improvement over existing state-of-the-art methods with respect to the PSNR and SSIM metrics across all datasets. The original inputs, the ground truth, and the qualitative results for Rain200H are shown in Figure \ref{fig:results rain200h}. As is evident from Figure \ref{fig:results rain200h}, JORDER \cite{yang2017deep}, RESCAN \cite{li2018recurrent}, and SPANet \cite{wang2019spatial} fail to remove the rain streaks from the heavy rain images. PReNet \cite{ren2019progressive}, ReHEN \cite{yang2019single}, RCDNet \cite{wang2020model}, and MPRNet \cite{zamir2021multi} remove almost all the rain streaks from the heavy rain images, but they fail to reconstruct the details of the background. On the other hand, it is apparent that MCW-Net successfully removes all the rain streaks and reconstruct the background more effectively than the other state-of-the-art methods. In particular, the digits of the plane are clearly restored in the first row of Figure \ref{fig:results rain200h}.}

%table results
As mentioned in Section \ref{sec:exp}, the proposed MCW-Net is evaluated on four synthetic datasets, and the performance is compared to eight state-of-the-art methods. The quantitative results on the synthetic datasets are presented in Table \ref{table:syn}.

For other models to be compared, if a metric is not provided in the original paper, we train the models with their default settings and report the results to the comparison table.  Otherwise, we report the better result between the provided metric in the original paper and the result obtained by our re-trained model. If reproduction is not possible, we directly copy the provided result to the comparison table.

As is evident from the data, the proposed MCW-Net (large) achieves remarkable improvement over existing state-of-the-art methods with respect to the PSNR and SSIM metrics across all synthetic datasets, and MCW-Net (small) follows right behind. 

%figure results
The original inputs, the ground truth, and the qualitative results for Rain200H are shown in Figure \ref{fig:results rain200h}. As shown in the yellow boxes of Figure \ref{fig:results rain200h}, MCW-Net (small) clearly restores the number compared to other methods, and MCW-Net (large) restores the digits surprisingly similar to ground truth. In the red boxes of Figure \ref{fig:results rain200h}, MCW-Net (small) restores the sky and spokes of the windmill cleanly compared to other methods but failed to recover lines, while MCW-Net (large) restores some of the lines.

\subsubsection{Results on Real-world Datasets} 
%about SPA-Data

% \textcolor{blue}{
% For further general verification of the proposed method, additional experiments are conducted on two real-world datasets \cite{wang2019spatial,yang2017deep}. On the SPA-Data \cite{wang2019spatial}, MCW-Net exhibits quantitatively superior performance compared to the other state-of-the-art methods \cite{deng2020detail, li2018recurrent, ren2019progressive, wang2020model,wang2019spatial, yang2017deep,yang2019single}. The qualitative evaluation is based on the degree to which the rain streaks are removed and the quality of the restored background, as in Section \ref{subsec:results_synthetic}. Most methods \cite{deng2020detail, ren2019progressive, wang2020model, wang2019spatial, yang2017deep} are incapable of completely erasing the actual rain, but the images obtained via RCDNet \cite{wang2020model}, ReHEN \cite{yang2019single} and MCW-Net are almost identical to the ground-truth, as shown in Figure \ref{fig:results SPA-Data}.}

% %about real-world data
% \textcolor{blue}{
% To confirm the effectiveness of the method trained using synthetic rainy images in removing real rain streaks, qualitative experiments are conducted on real-world rainy images. To compare the model performances under fair conditions, only Rain200H \cite{yang2017deep} is used during the training process. As shown in Figure \ref{fig:results real}, MCW-Net generates satisfactory results with respect to both the removal of the rain streaks and the restoration of the details in the background.}

For further general verification of the proposed method, additional experiments are conducted on two real-world datasets. To estimate the performance of the other state-of-the-art models, we employ the same way as quantitative evaluation of synthetic data. On the SPA-Data, MCW-Net exhibits quantitatively outstanding performance compared to the other state-of-the-art methods.

%about real-world data
To confirm the effectiveness of the method trained using synthetic rainy images in removing real rain streaks, qualitative experiments are conducted on images presented by Yang \textit{et al.} \cite{yang2017deep}. To compare the model performances under fair conditions, only Rain200H is used during the training process. As shown in Figure \ref{fig:results real}, MCW-Net generates satisfactory results with respect to both the removal of the rain streaks and the restoration of the details in the background. The small and large versions of MCW-Net recover the details of the columns in the red box and remove the rain streaks in the yellow box better compared to other models. Although detail recovery and rain removal are the trade-off for other models, MCW-Net succeeds in both. MCW-Net also recovers the cleanest background for another image sample. The yellow box shows the output of MCW-Net exhibits no traces of rain streaks while they are left in the result of the other models.

\subsubsection{Results on Authentic Rain Images with the Dedicated Metric}
\label{subsec:results_authentic}

PSNR and SSIM estimate how the recovered output image closes to the target image. Therefore, several low-level vision tasks (\eg, denoising, super-resolution, deraining) conventionally exploit these metrics as an evaluation tool. Nonetheless, one might argue that PSNR and SSIM are general-purpose quality metrics so they are limited to concentrating only on the deraining ability. Besides, they need target images to be calculated and thus cannot be applied on target-absent authentic rainy images.\\
To handle this issue, we additionally evaluate the proposed method via a measurement called B-FEN~\cite{wu2020subjective}, which accurately evaluates the deraining quality using a bi-directional feature embedding network. A higher B-FEN score represents better perceptual quality, which indicates that the model not only effectively removes rain streaks but also well preserves the original rain-free image. We train all the comparable models and the proposed method on SPA-Data and evaluate the DQA dataset~\cite{wu2020subjective}. Since DQA is a real-world testing image set, real-world SPA-Data can guide the model to capture the properties of authentic rain streaks.\\
As shown in Table \ref{bfen}, our model achieves the highest B-FEN score. One thing to note is that the small version of MCW-Net has a higher B-FEN score than the large version of MCW-Net. B-FEN is a subjective opinion-aware metric, and opinion-making participants may tend to focus more on rain streaks removal than background restoration. From this point of view, it may lead to possible inconsistent results different from that of other objective opinion-unaware metrics.

\begin{table}[]
	\tabcolsep 0.15in{\scriptsize{}}
	\centering
	\caption{Comparison results of the various methods on DQA dataset in B-FEN \cite{wu2020subjective} metric (higher is better).  }
	\vspace{0.15cm}
	
	\begin{tabular*}{\tblwidth}{@{}LL@{}}
		\toprule
		Methods                    & B-FEN  \\%& NIQE   
		\midrule
		original                 & 0.2997 \\%& 14.7787 \\ 
		MPRNet            & 0.3051 \\%&  14.0417\\ 
		RCDNet            & 0.3101 \\%&  14.5701\\ 
		PreNet            & 0.3139 \\%& 13.8142 \\ 
		SPANet            & 0.3154 \\%& 15.0969 \\ 
		MCW-Net (ours-small)   & 0.3287 \\%& 14.0555 \\ 
		MCW-Net (ours-large)   & 0.3222 \\ % & 13.7932 \\ 
		\bottomrule
	\end{tabular*}
	
	\smallskip
	\label{bfen}
\end{table}

\subsubsection{Results on Raindrop data}
\label{subsec:results_raindrop}

\begin{figure*}
	\centering
	\setlength{\tabcolsep}{0pt}
	\footnotesize{
	\begin{tabular}{cccccccclcccccccclcccccccclcccccccclccccccccl}
		\multicolumn{3}{c}{\includegraphics[width=0.19\textwidth]{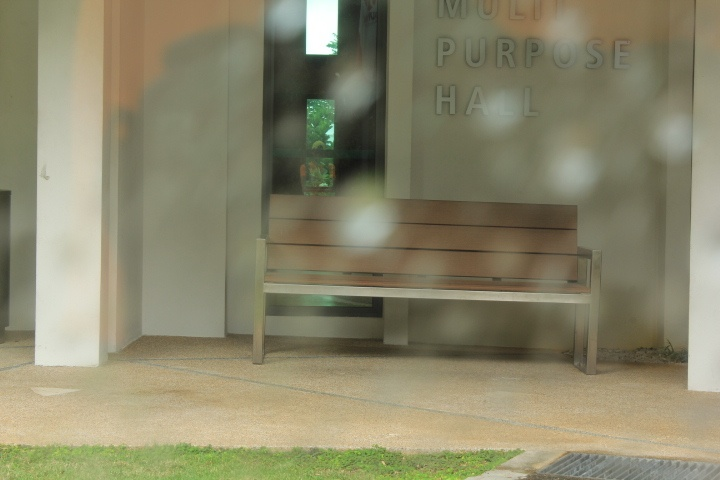}}\ &
		\multicolumn{3}{c}{\includegraphics[width=0.19\textwidth]{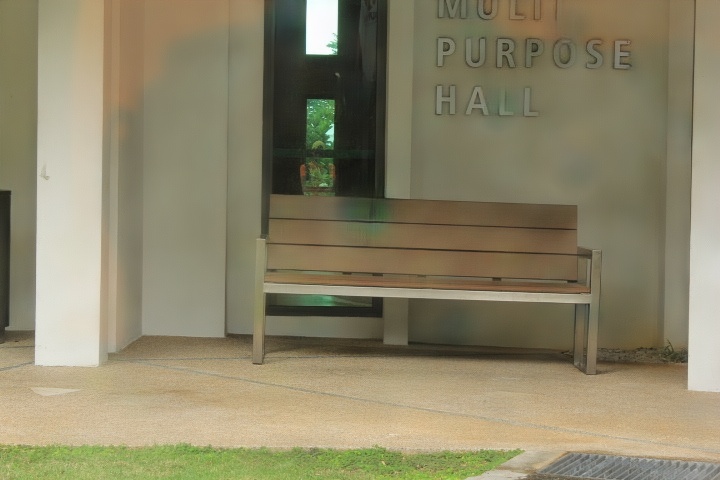}}\ &
		\multicolumn{3}{c}{\includegraphics[width=0.19\textwidth]{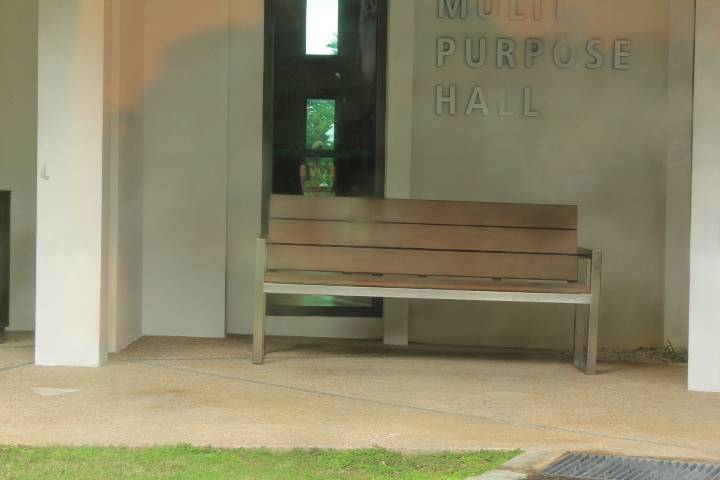}}\ &
		\multicolumn{3}{c}{\includegraphics[width=0.19\textwidth]{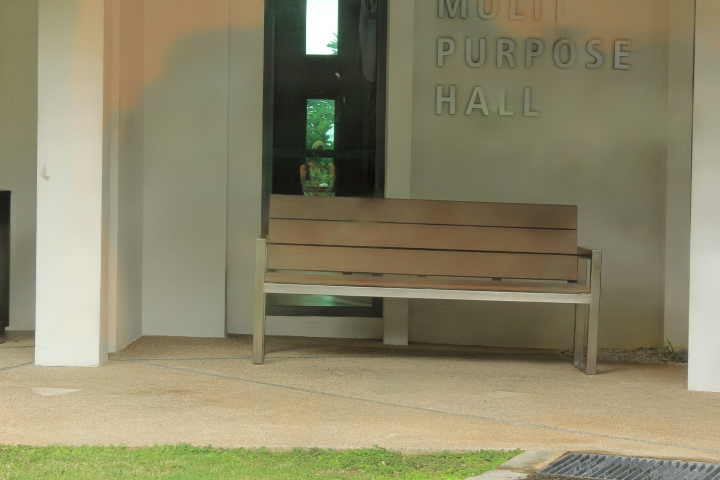}}\ &
		\multicolumn{3}{c}{\includegraphics[width=0.19\textwidth]{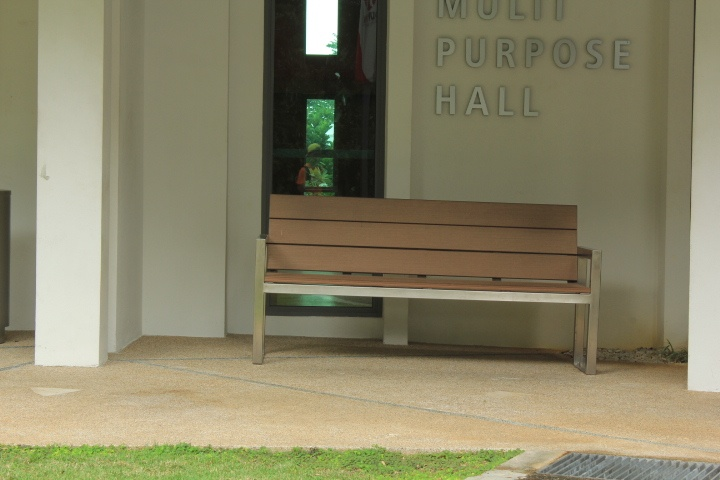}}\\
		
		\multicolumn{3}{c}{(a) Rain drop image} &
		\multicolumn{3}{c}{(b) AGAN~\cite{qian2018attentive}} &
		\multicolumn{3}{c}{(c) MCW-Net (small)} &
		\multicolumn{3}{c}{(d) MCW-Net (large)} &
		\multicolumn{3}{c}{(e) GT} \\

		\multicolumn{3}{c}{\includegraphics[width=0.19\textwidth]{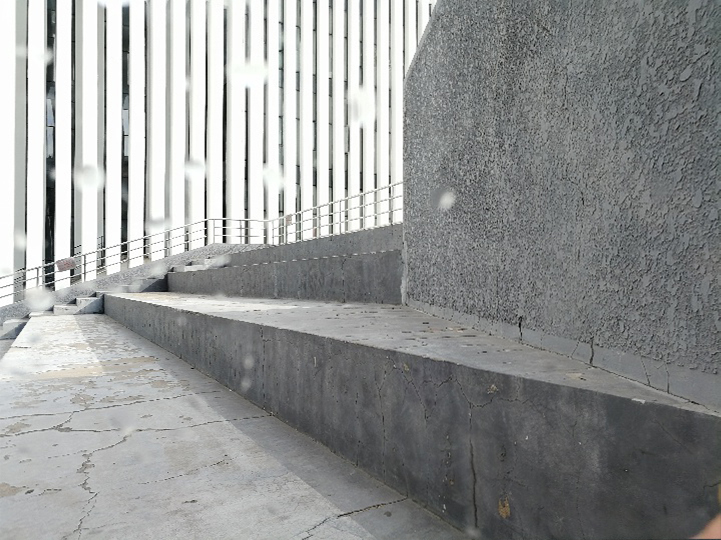}}\ &
		\multicolumn{3}{c}{\includegraphics[width=0.19\textwidth]{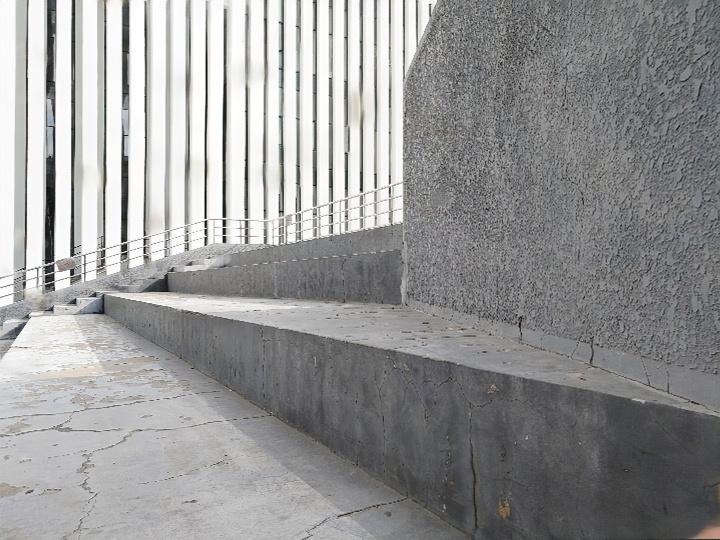}}\ &
		\multicolumn{3}{c}{\includegraphics[width=0.19\textwidth]{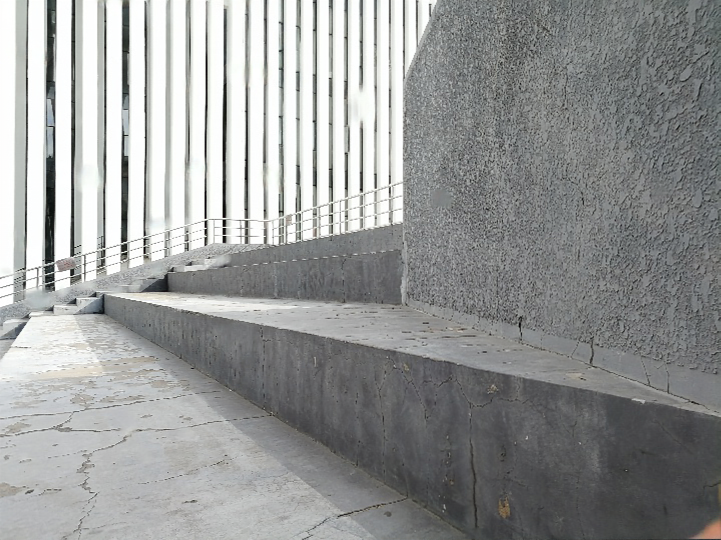}}\ &
		\multicolumn{3}{c}{\includegraphics[width=0.19\textwidth]{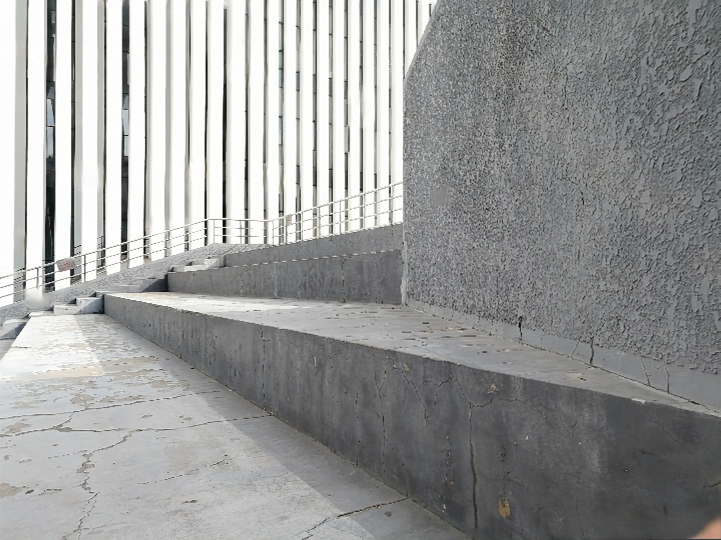}}\ &
		\multicolumn{3}{c}{\includegraphics[width=0.19\textwidth]{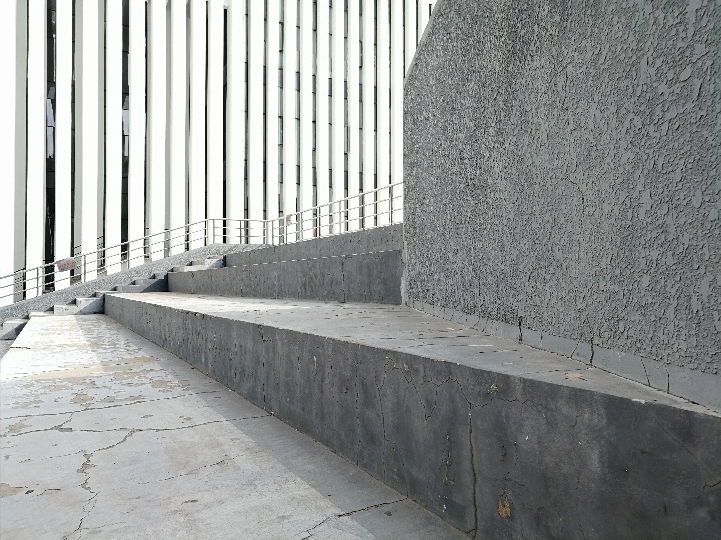}}\\
		
		\multicolumn{3}{c}{(a) Rain drop image} &
		\multicolumn{3}{c}{(b) AGAN~\cite{qian2018attentive}} &
		\multicolumn{3}{c}{(c) MCW-Net (small)} &
		\multicolumn{3}{c}{(d) MCW-Net (large)} &
		\multicolumn{3}{c}{(e) Clean} \\

	\end{tabular}}
	\caption{ Results obtained via several state-of-the-art methods on the Raindrop images. Images in the first ans second rows are from testset A and testset B, respectively. }
	\label{fig:results raindrop}
\end{figure*}

Raindrops, which are a commonly observed phenomenon in conjunction with rain, also might degrade the performance in computer vision applications. Even though we design the proposed method to remove rain streaks in images, we explore the model's generalizability with the raindrop image dataset. The experimental results are reported in Table \ref{raindrop} and Figure \ref{fig:results raindrop}. In the evaluation, we use the weight of the AGAN model provided by the author. We calculate PSNR and SSIM metrics in RGB channels as in other experiments. 

\begin{table}[]
	\tabcolsep 0.15in{\scriptsize{}}
	\centering
	\caption{ Average PSNR and SSIM comparision on Raindrop dataset. }
	\vspace{0.15cm}
	
	\begin{tabular*}{\tblwidth}{@{}LLLLL@{}}
		\toprule
		Dataset                        & \multicolumn{2}{c}{Testset A} &  \multicolumn{2}{c}{Testset B} \\
		                               & PSNR  & SSIM & PSNR  &  SSIM  \\ \midrule
   		Eigen13 \cite{eigen2013restoring}  & 23.74 & 0.788 & - & -\\ 
  		Pix2Pix \cite{isola2017image}  & 28.15 & 0.855 & - & -\\ 
  		PreNet \cite{ren2019progressive}  & 28.58 & 0.913 & - & -\\ 
		AGAN \cite{qian2018attentive}  & 30.55 & 0.910 & 24.43 & 0.795\\ 
		MCW-Net (ours-small)               & 29.96 & 0.906 & 24.91 & 0.800 \\ 
		MCW-Net (ours-large)               & 30.77 & 0.918 & 25.17 & 0.809  \\ \bottomrule
	\end{tabular*}
	
	\smallskip
	\label{raindrop}
\end{table}

\begin{table}[]
	\tabcolsep 0.15in{\scriptsize{}}
	\centering
	\caption{Ablation study on types of regional non-local blocks. }
	\vspace{0.15cm}
	
	\begin{tabular*}{\tblwidth}{@{}LLLL@{}}
		\toprule
		Dataset & Region Type                    & PSNR  & SSIM   \\ \midrule
		         & Tall Rectangle                 & 30.08 & 0.916 \\ 
		Rain200H &Square                          & 30.14 & 0.915 \\ 
		         & Wide Rectangle                 & 30.62 & 0.921 \\
		\midrule
		         & Tall Rectangle                 & 39.86 & 0.987 \\ 
		Rain200L &Square                          & 39.87 & 0.988 \\ 
		         & Wide Rectangle                 & 39.92 & 0.988 \\
		\midrule
                 & Tall Rectangle                 & 42.78 & 0.987 \\ 
		SPA-DATA &Square                          & 42.96 & 0.987 \\ 
                 & Wide Rectangle                 & 43.05 & 0.987 \\
		         \bottomrule
	\end{tabular*}
	\smallskip
	\label{nl ablation}
\end{table}

\begin{table}[]
	\centering
	\caption{ Ablation study on the various strategies presented in Section \ref{sec:proposed}.}
	\vspace{0.15cm}
	{\small
	\begin{tabular*}{\tblwidth}{@{}LLLLLL@{}}
		\toprule
		
		WRNL        &       DWT       & MLC                                & Cutmix        & PSNR     & SSIM   \\ \midrule
	                &                 &                                    &               & $28.12$  & $0.906$\\ 
		\checkmark  &                 &                                    &               & $29.49$  & $0.911$\\           
		\checkmark  &   \checkmark    &                                    &               & $30.22$  & $0.917$\\  
% 		\checkmark  &   \checkmark    & \checkmark                         &               & $29.07$  & $0.895$\\  
		\checkmark  &   \checkmark    & \checkmark                         &               & $30.62$  & $0.921$\\ 
		\checkmark  &   \checkmark    & \checkmark                         &  \checkmark   & $30.70$  &  $0.922$\\ \bottomrule
	\end{tabular*}
	}
% 	\smallskip
	\label{total ablation}
\end{table}

\begin{table}
	\caption{Ablation study on various sampling methods. Note that three different sampling operations are compared on the proposed method without MLP and Cutmix.}
	
	\begin{tabular*}{\tblwidth}{@{}LLL@{}}
		\toprule  
		Sampling Operation               & PSNR  & SSIM  \\ 
		\midrule
		Mean Pooling                     & 29.50 & 0.909 \\ 
		1$\times$1 conv.                 & 29.80 & 0.911 \\
	    DWT $\&$ IWT                     & 30.62 & 0.921 \\ 
	    \bottomrule
	\end{tabular*} 
	\label{tab:sampling-ablation}
\end{table}

\subsection{Ablation Study}
% ychpark_see

We conduct an ablation study to demonstrate the significance of all the methods used in the MCW-Net architecture. MCW-Net (large) and Rain200H dataset are used for the ablation study. We conduct three experiments and report the average values. All the evaluations are dedicated to the proposed method without Cutmix. Because Cutmix is the data augmentation strategy and hence is not directly related to ablation about the model structure.

\begin{table}
	\caption{Ablation study on MLC, where C.A. denotes channel-wise attention. The experiments are conducted on the proposed method without Cutmix.}
	
	\begin{tabular*}{\tblwidth}{@{}LLL@{}}
		\toprule  
		                               & PSNR  & SSIM  \\ 
		\midrule
		No MLC                          & 30.22 & 0.917\\
		MLC with concatenation       & 30.07 & 0.913 \\
		MLC with addition            & 30.26 & 0.916 \\
		MLC with C.A. (SE)           & 30.62 & 0.921 \\
	    \bottomrule
	\end{tabular*}
	\label{tab:mlc-ablation}
\end{table}

\subsubsection{Ablation study on strategies employed} 
An ablation investigation is conducted to evaluate the performance of the proposed strategies. The baseline model is constructed with two DCR blocks corresponding to each stage and the 2$\times$2 max pooling and pixel shuffle operation are adopted as the down-sampling and up-sampling operations, respectively. As evident from Table \ref{total ablation}, each strategy contributes to the performance improvement.
% In particular, applying MLC without the SE block can degrade the performance. Leveraging SE block to consider attentions between the levels makes MLC effective on the model.

\begin{figure*}[!t]\footnotesize
	\centering
	\setlength{\tabcolsep}{0pt}
	
	\begin{tabular}{cclcclcclccl}
	    \multicolumn{3}{c}{\includegraphics[width=.25\textwidth]{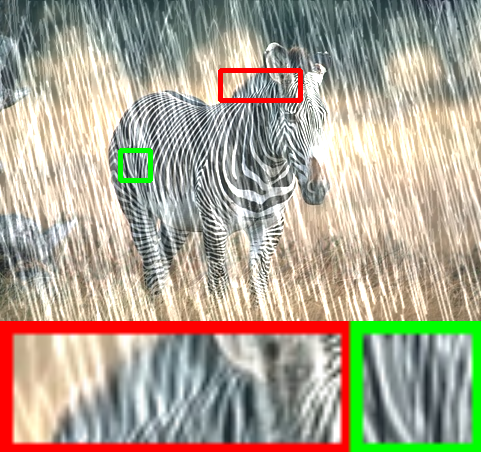}} &
		\multicolumn{3}{c}{\includegraphics[width=.25\textwidth]{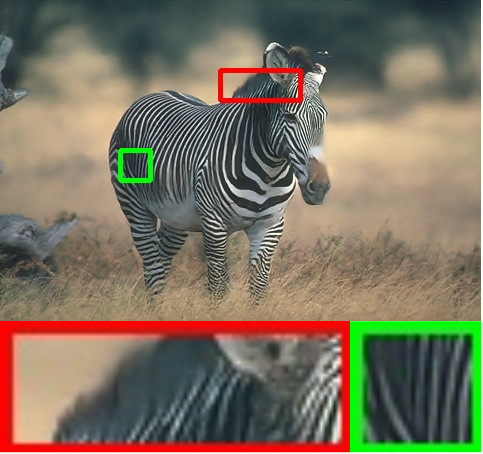}} &
		\multicolumn{3}{c}{\includegraphics[width=.25\textwidth]{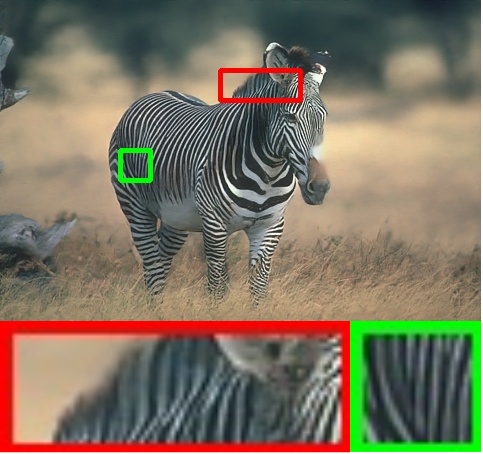}} &
		\multicolumn{3}{c}{\includegraphics[width=.25\textwidth]{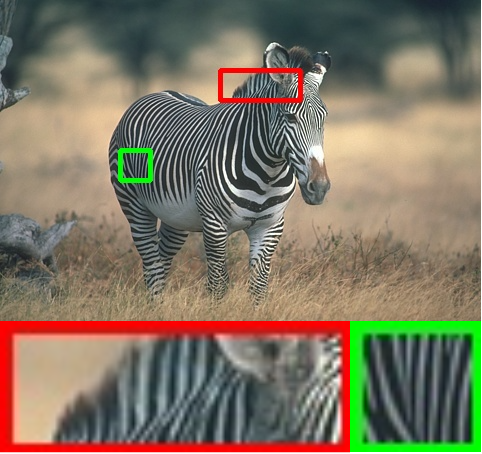}}
		\\
		\multicolumn{3}{c}{\includegraphics[width=.25\textwidth]{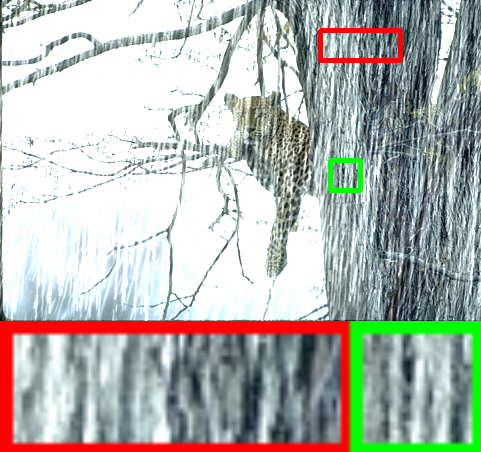}} &
		\multicolumn{3}{c}{\includegraphics[width=.25\textwidth]{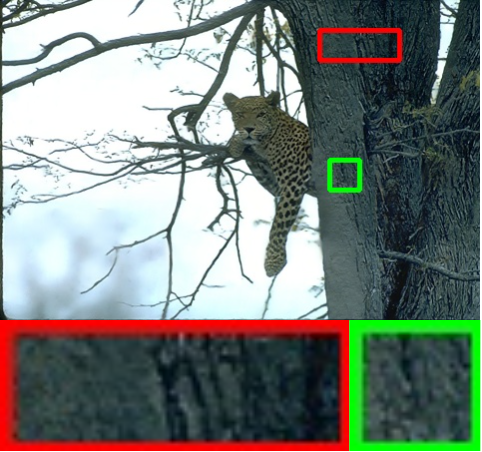}} &
		\multicolumn{3}{c}{\includegraphics[width=.25\textwidth]{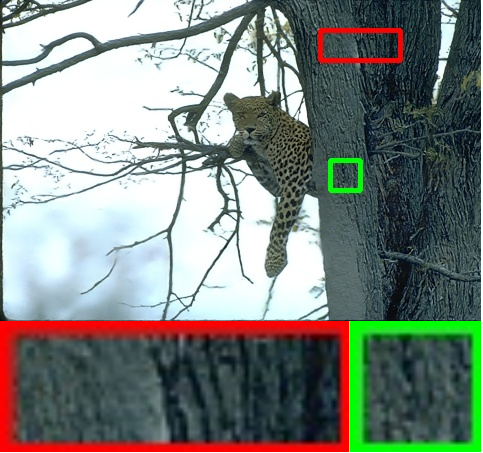}} &
		\multicolumn{3}{c}{\includegraphics[width=.25\textwidth]{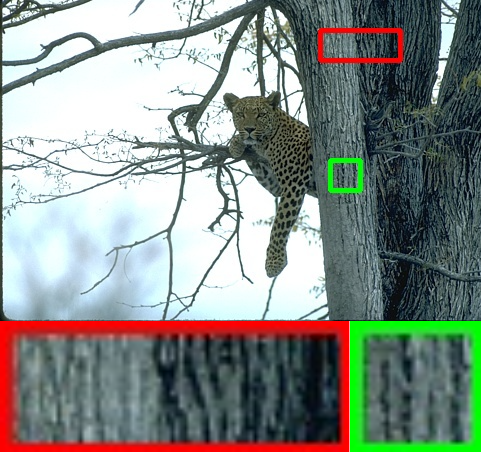}}
		\\
		\multicolumn{3}{c}{(a) {Input}} &
		\multicolumn{3}{c}{(b) {MCW-Net (w.o. MLC)}} &
		\multicolumn{3}{c}{(c) {MCW-Net (w. MLC)}} &
		\multicolumn{3}{c}{(d) {GT}}
	\end{tabular}
	\caption{Qualitative ablation study on MLC. MCW-Net (large) is used for the study. In the first row, we can see that the zebra pattern in the red and green box is not well restored without MLC. However, model with MLC restored the pattern in the red and green box well. In the second row, we can also see that the model with MLC better restored the tone and texture of the tree than the model without MLC. As a result, we can confirm that MLC plays a certain role in recovering detail as intended.}

	\label{fig:mlc_ablation}
\end{figure*}

\subsubsection{Ablation study on MLC}
\label{sec:mlc_ablation}
% To show the importance of channel-wise attention to MLC, we evaluate the performance using MLC with channel-wise attention and MLC without channel-wise attention. 
To show the importance of channel-wise attention to MLC, we evaluate the performance using MLC with channel-wise attention and MLC with other commonly used fusing operations, addition and concatenation.
As shown in Table \ref{tab:mlc-ablation}, MLC with addition or concatenation rather degrade the performance. We analyze that this result occurs because additional information which is messy and unorganized rather interferes with the decoding process. Considering channel-wise attention serves as an indication of which information should be referenced more importantly at the current level decoding process, so MLC with channel-wise attention improves the performance of the model. 
% \red{For instance, stage 4 in Figure~\ref{} necessitates more high-level features (\textit{e.g.}, entire object with significant variation) than the other-level features and stage 7 needs more low-level features (\textit{e.g.}, corners or edge/color conjunctions) for better recovering.}

In addition, we conduct a qualitative ablation study to see if MLC actually helps to restore the details as intended, and the results are shown in Figure \ref{fig:mlc_ablation}. The results confirm that MLC effectively does detail recovery as well as quantitative improvement. Details of the results are described in the caption of Figure \ref{fig:mlc_ablation}.

\subsubsection{Ablation study on non-local block region types} 
We evaluate the performance using square, tall, and wide-type regional non-local blocks on Rain200H, Rain200L, and SPA-DATA datasets. The results presented in Table \ref{nl ablation} demonstrate that the wide-type regional non-local block achieves the best performance.

\subsubsection{Ablation study on various sampling operations} 
To compare the performance of various sampling operations, we evaluate the performance using mean pooling, 1x1 convolution, and the discrete wavelet transform. The results presented in Table \ref{tab:sampling-ablation} demonstrate that the discrete wavelet transform has the best performance.

\begin{table*}[]
	\centering
	\caption{ Comparison results of joint deraining and semantic segmentation on RainCityscape dataset comprising three rain intensities ($\alpha \in \{0.01,0.02,0.03 \}$ where $\alpha$ denotes the intensity of the rain streaks). We use DeepLabV3+ \cite{chen2018encoder} for semantic segmentation. We compare the models that show an improvement in the semantic segmentation performance which is measured as mIOU metric. avg. in the metric column denotes average value of all $\alpha$.
	}
	\begin{tabular*}{\tblwidth}{@{}LLLLLLLLL@{}}
		\toprule
		Metric  &  Rainy   &   ReHEN    & PReNet                      & RCDNet      &  MPR-Net &  MCW-Net  &  MCW-Net (Large) &  Clean   \\ \midrule
		                                  %&   (MM' 2019)     & (CVPR'19)                   & (CVPR'20)      &     (our)   &   (No rain)                \\
		
		PSNR                   &  15.55  & 23.47 & $28.88$                           &   $25.51$     &25.91           &   $\bold{33.94}$  & $\bold{35.82}$  &  $\infty$   \\ 
		SSIM                   &  0.826  & 0.916 & $0.972$                           &   $0.958$        &   0.964          &   $\bold{0.981}$  & $\bold{0.987}$    &  $1.000$   \\ 
		                  %&    &                         &                    &         & \\ 

		mIOU (avg.)                &  0.6254 & 0.4833 & $0.7636$                         &  $0.7402$                   &  0.7516       &   $\bold{0.7679}$  & $\bold{0.7728}$     & $0.7810$\\ 
		mIOU ($\alpha$=$0.01$)                  & 0.5528 &  0.6724  & $0.7765$                         &  $0.7641$        & 0.7626              &   $\bold{0.7743}$  & $\bold{0.7773}$     &  \\ 
		mIOU ($\alpha$=$0.02$)                  & 0.4816 &  0.6284  & $0.7652$                         &  $0.7439$        & 0.7509            &   $\bold{0.7703}$   & $\bold{0.7750}$   &  \\ 
		mIOU ($\alpha$=$0.03$)                  & 0.4171 &  0.5777  & $0.7492$                         &  $0.7140$        & 0.7414            &   $\bold{0.7590}$  & $\bold{0.7663}$      &  \\   \bottomrule
		
\end{tabular*}
\label{table:cityscape}

\end{table*}

\begin{figure*}[]\footnotesize
	\centering
	\setlength{\tabcolsep}{0pt}
	\begin{tabular}{cclcclcclcclcclcclcclcclcclCCL}
		\multicolumn{3}{c}{\includegraphics[width=0.14\textwidth]{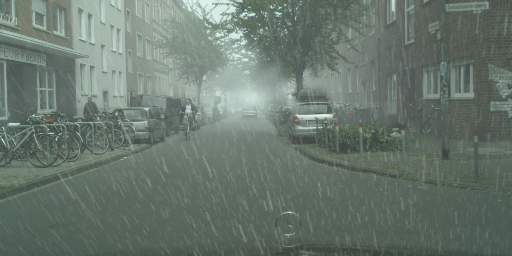}}\ &
		\multicolumn{3}{c}{\includegraphics[width=0.14\textwidth]{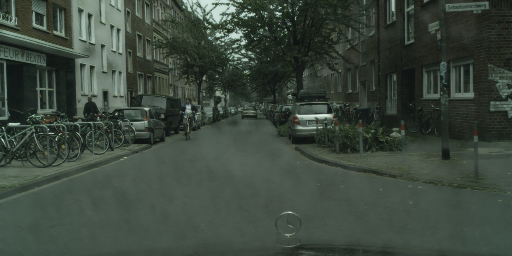}}\ &
		\multicolumn{3}{c}{\includegraphics[width=0.14\textwidth]{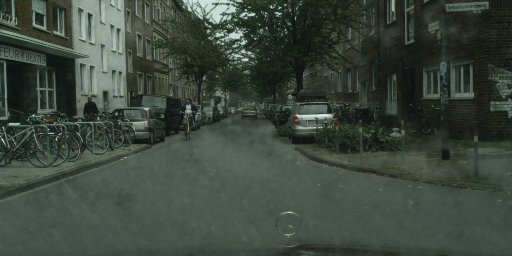}}\ &
		\multicolumn{3}{c}{\includegraphics[width=0.14\textwidth]{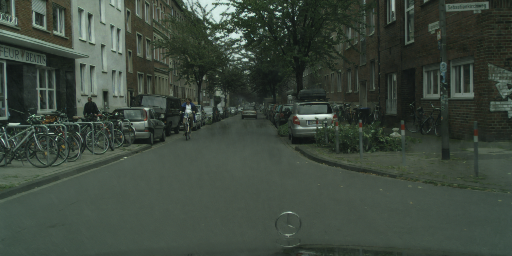}}\ &
		\multicolumn{3}{c}{\includegraphics[width=0.14\textwidth]{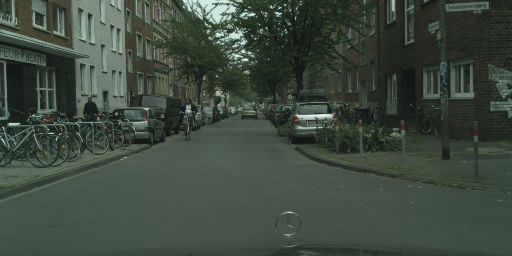}}\ &
		\multicolumn{3}{c}{\includegraphics[width=0.14\textwidth]{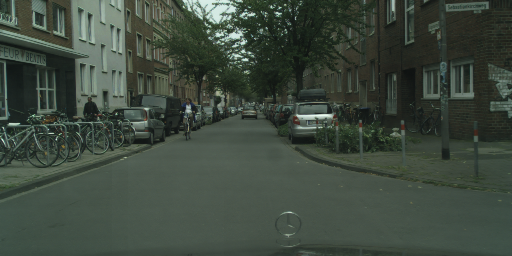}}\ &
		\multicolumn{3}{c}{\includegraphics[width=0.14\textwidth]{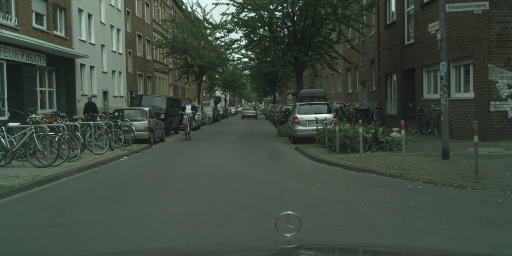}}\\
		
		\multicolumn{3}{c}{\includegraphics[width=0.14\textwidth]{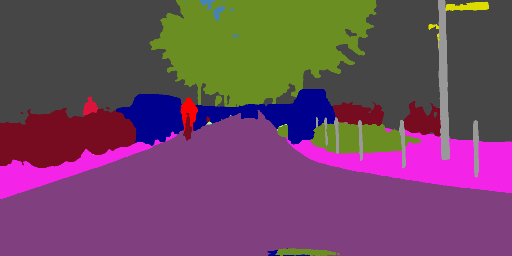}}\ &
		\multicolumn{3}{c}{\includegraphics[width=0.14\textwidth]{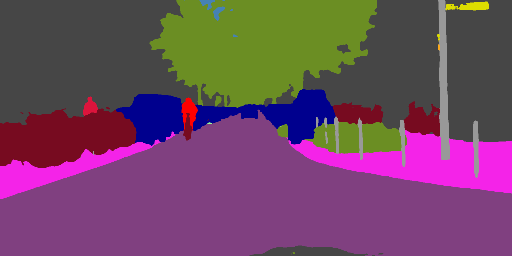}}\ &
		\multicolumn{3}{c}{\includegraphics[width=0.14\textwidth]{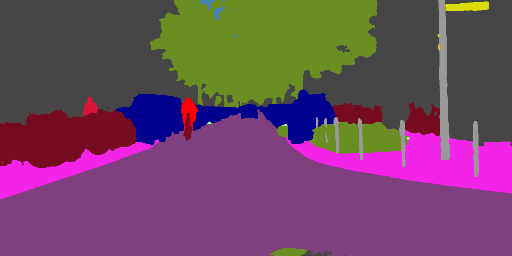}}\ &
		\multicolumn{3}{c}{\includegraphics[width=0.14\textwidth]{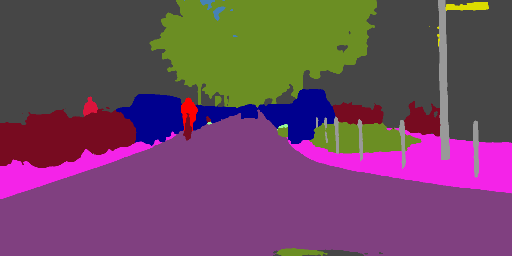}}\ &
		\multicolumn{3}{c}{\includegraphics[width=0.14\textwidth]{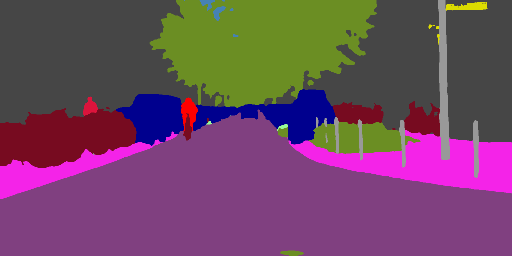}}\ &
		\multicolumn{3}{c}{\includegraphics[width=0.14\textwidth]{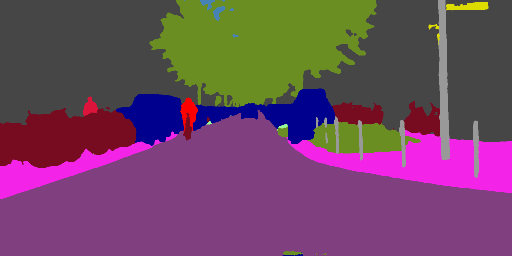}}\ &
		\multicolumn{3}{c}{\includegraphics[width=0.14\textwidth]{figure/RainCityscape/lmunster_000015_norain_seg.png}}\\
		
		\multicolumn{3}{c}{\includegraphics[width=.14\textwidth]{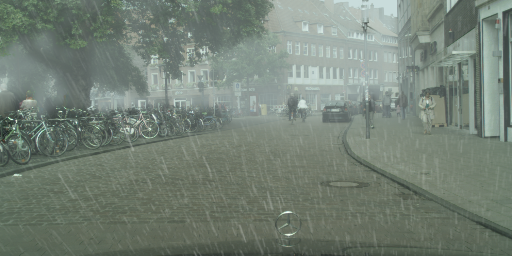}}\ &
		\multicolumn{3}{c}{\includegraphics[width=.14\textwidth]{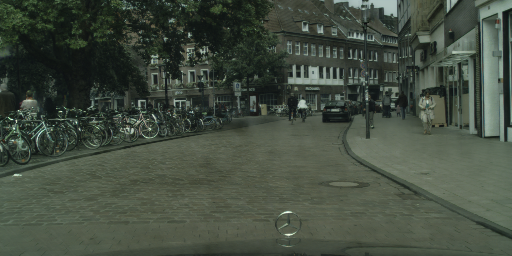}}\ &
		\multicolumn{3}{c}{\includegraphics[width=.14\textwidth]{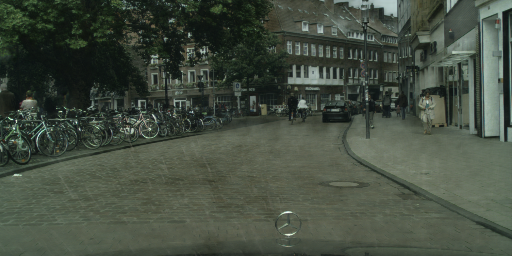}}\ &
		\multicolumn{3}{c}{\includegraphics[width=.14\textwidth]{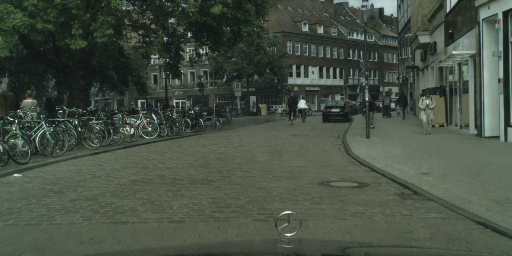}}\ &
		\multicolumn{3}{c}{\includegraphics[width=0.14\textwidth]{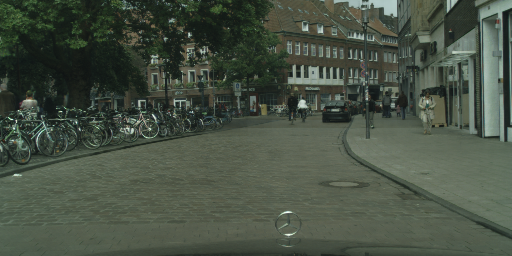}}\ &
		\multicolumn{3}{c}{\includegraphics[width=.14\textwidth]{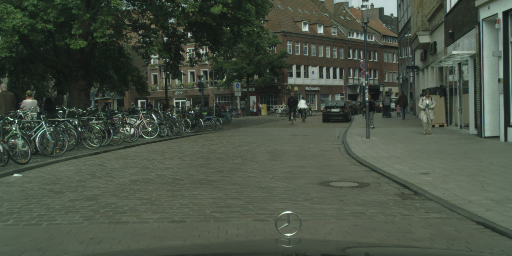}}\ &
		\multicolumn{3}{c}{\includegraphics[width=.14\textwidth]{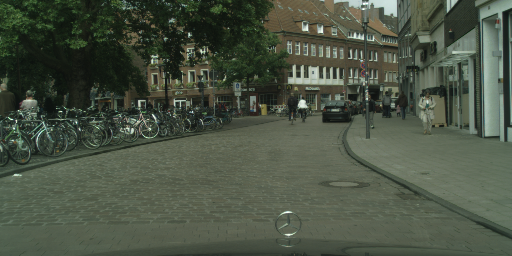}}\\
		
		\multicolumn{3}{c}{\includegraphics[width=.14\textwidth]{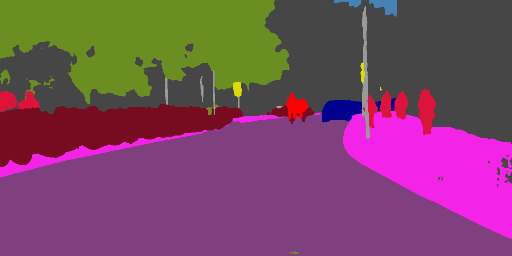}}\ &
		\multicolumn{3}{c}{\includegraphics[width=.14\textwidth]{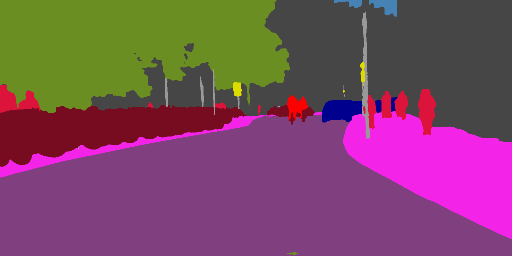}}\ &
		\multicolumn{3}{c}{\includegraphics[width=.14\textwidth]{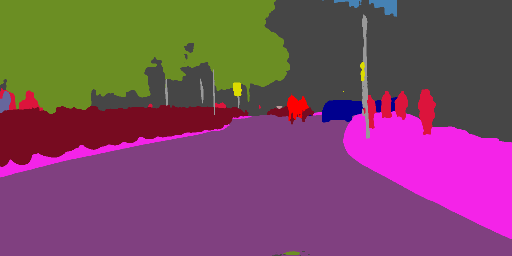}}\ &
		\multicolumn{3}{c}{\includegraphics[width=.14\textwidth]{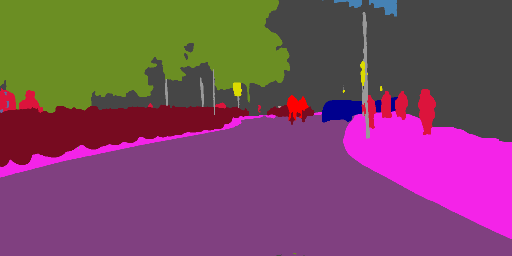}}\ &
		\multicolumn{3}{c}{\includegraphics[width=0.14\textwidth]{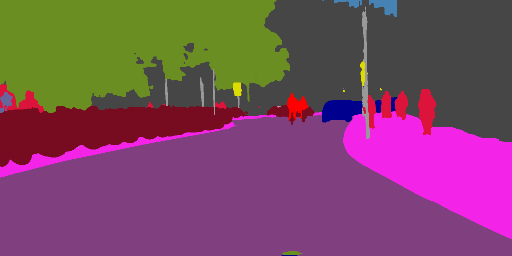}}\ &
		\multicolumn{3}{c}{\includegraphics[width=.14\textwidth]{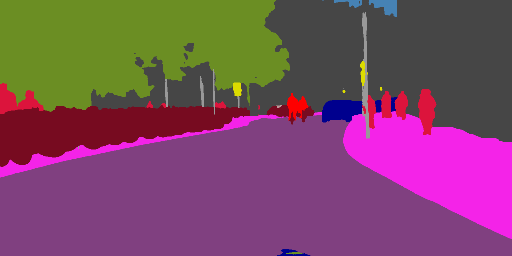}}\ &
		\multicolumn{3}{c}{\includegraphics[width=.14\textwidth]{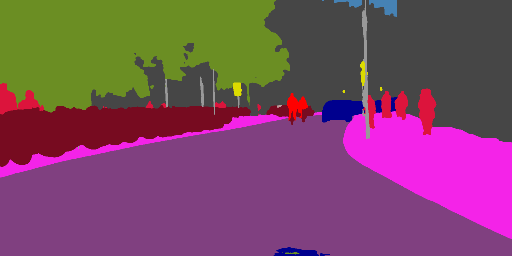}}\\

 		\multicolumn{3}{c}{(a) Rainy} &
		\multicolumn{3}{c}{(b) PReNet} &
		\multicolumn{3}{c}{(c) RCDNet} &
		\multicolumn{3}{c}{(d) MPR-Net} &
		\multicolumn{3}{c}{(e) Ours (small)} &
		\multicolumn{3}{c}{(f) Ours (large)} &
		\multicolumn{3}{c}{(g) Clean} 
	\end{tabular}
	\caption{Examples of joint deraining and semantic segmentation. The first row denotes the deraining results on the RainCityscape dataset. The second row denotes the semantic segmentation results obtained by DeepLabV3+ \cite{chen2018encoder}.}
	\label{fig:results cityscape}
\end{figure*}

\subsection{Applications for Other Tasks}
We investigate the effect of the deraining model on improving the performance of high-level vision applications such as semantic segmentation. Because rain streaks can degrade the visibility of objects under complex weather conditions, the incorporation of effective image enhancement would be helpful in several vision models. To this end, we apply the public semantic segmentation model DeepLabV3+ \cite{chen2018encoder} on the Cityscape dataset \cite{Cordts2016Cityscapes}. Hu \textit{et al.} \cite{hu2019depth} synthesized rain streaks on the Cityscape dataset with different rain intensities $\alpha$ ($\alpha \in \{0.01,0.02,0.03\} )$. Quantitative results for the improvement of the semantic segmentation accuracy in addition to the deraining performance are reported in Table \ref{table:cityscape}. The qualitative comparison is shown in Figure \ref{fig:results cityscape}.

\begin{figure*}[!h]\footnotesize
	\centering
	\setlength{\tabcolsep}{0pt}
	
	\begin{tabular}{cclcclccl}
		\multicolumn{3}{c}{\includegraphics[width=.4\textwidth]{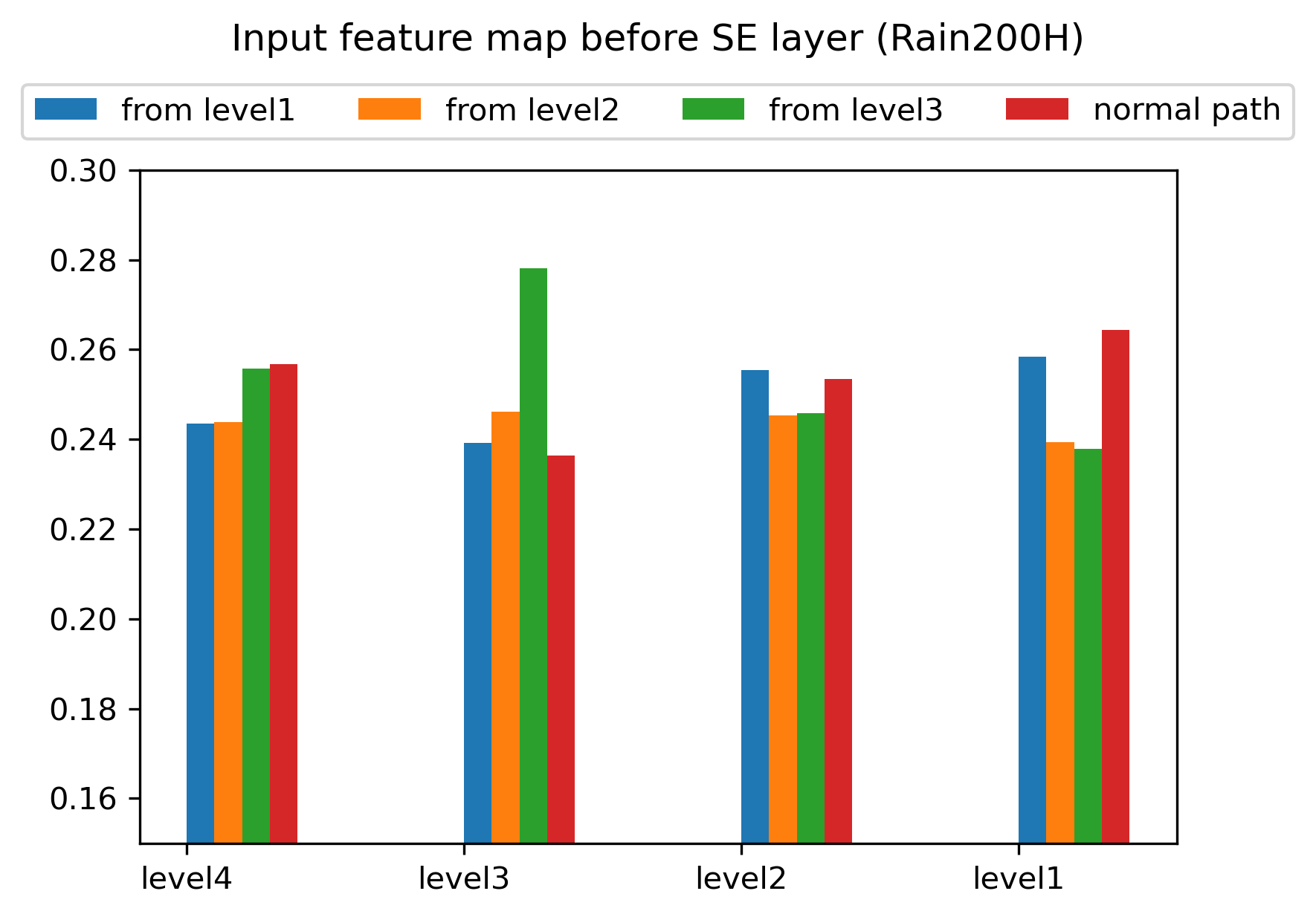}} &
		\multicolumn{3}{c}{\includegraphics[width=.4\textwidth]{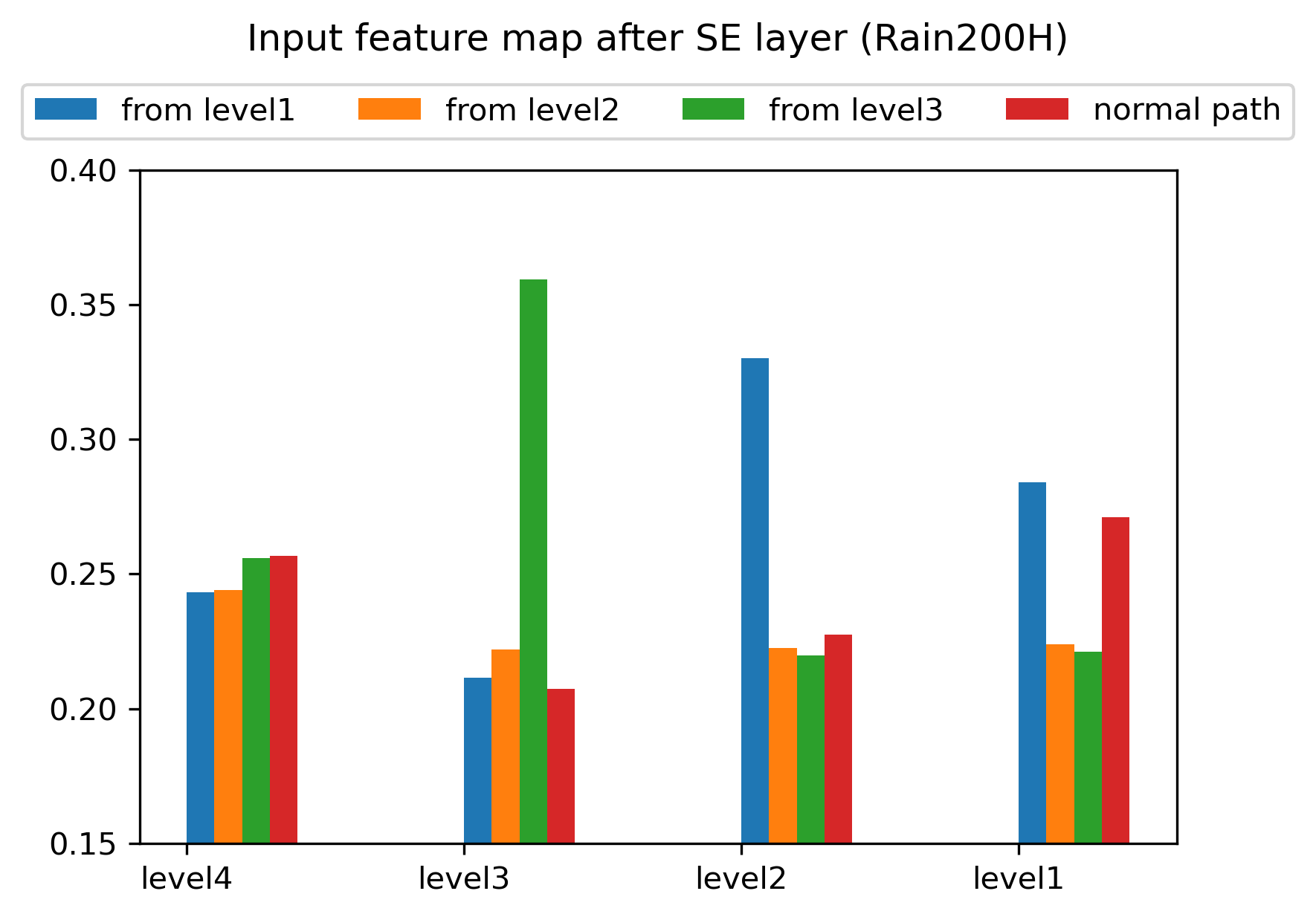}} \\
		\multicolumn{3}{c}{(a)} &
		\multicolumn{3}{c}{(b)} \\
		\multicolumn{3}{c}{\includegraphics[width=.4\textwidth]{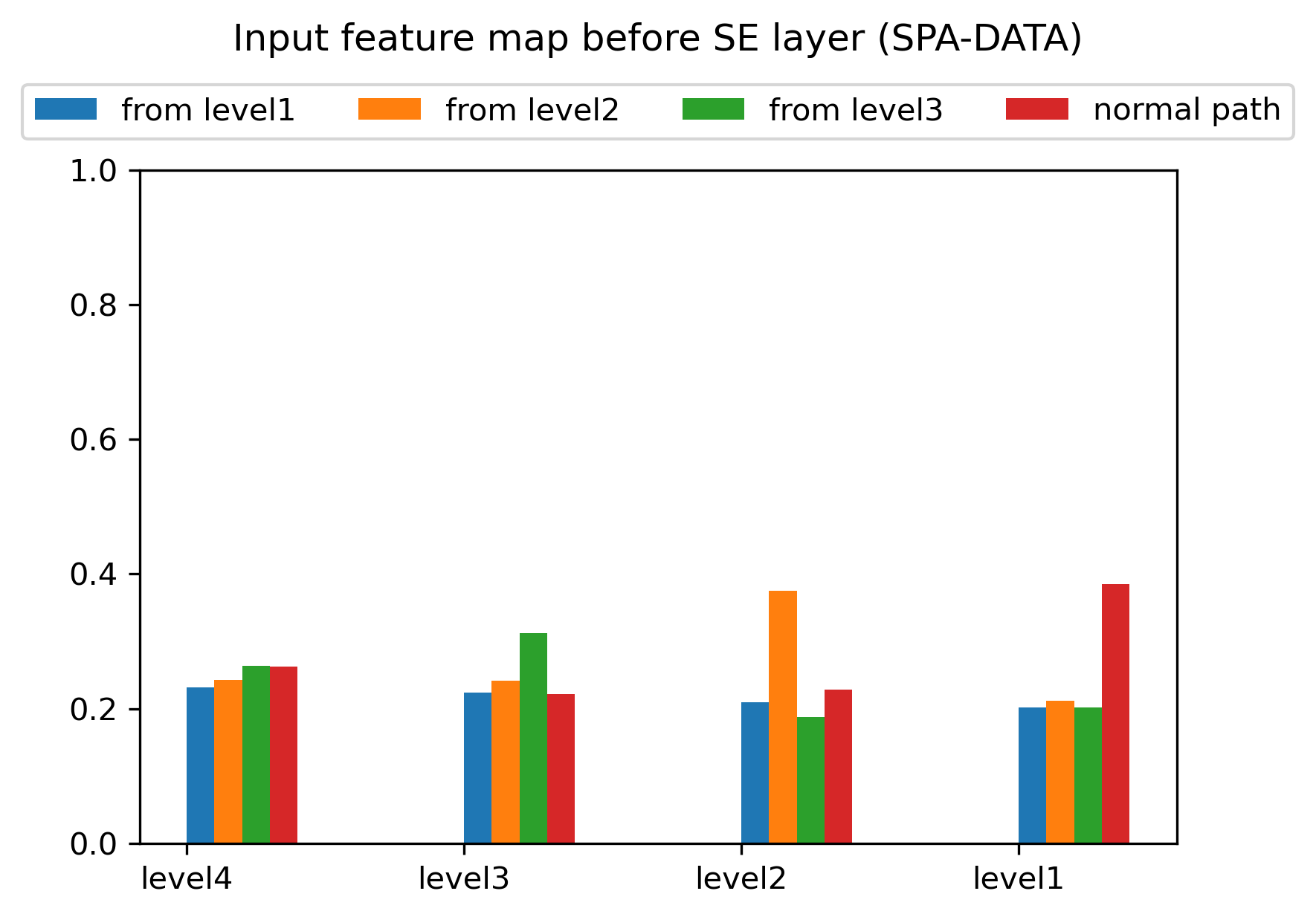}} &
		\multicolumn{3}{c}{\includegraphics[width=.4\textwidth]{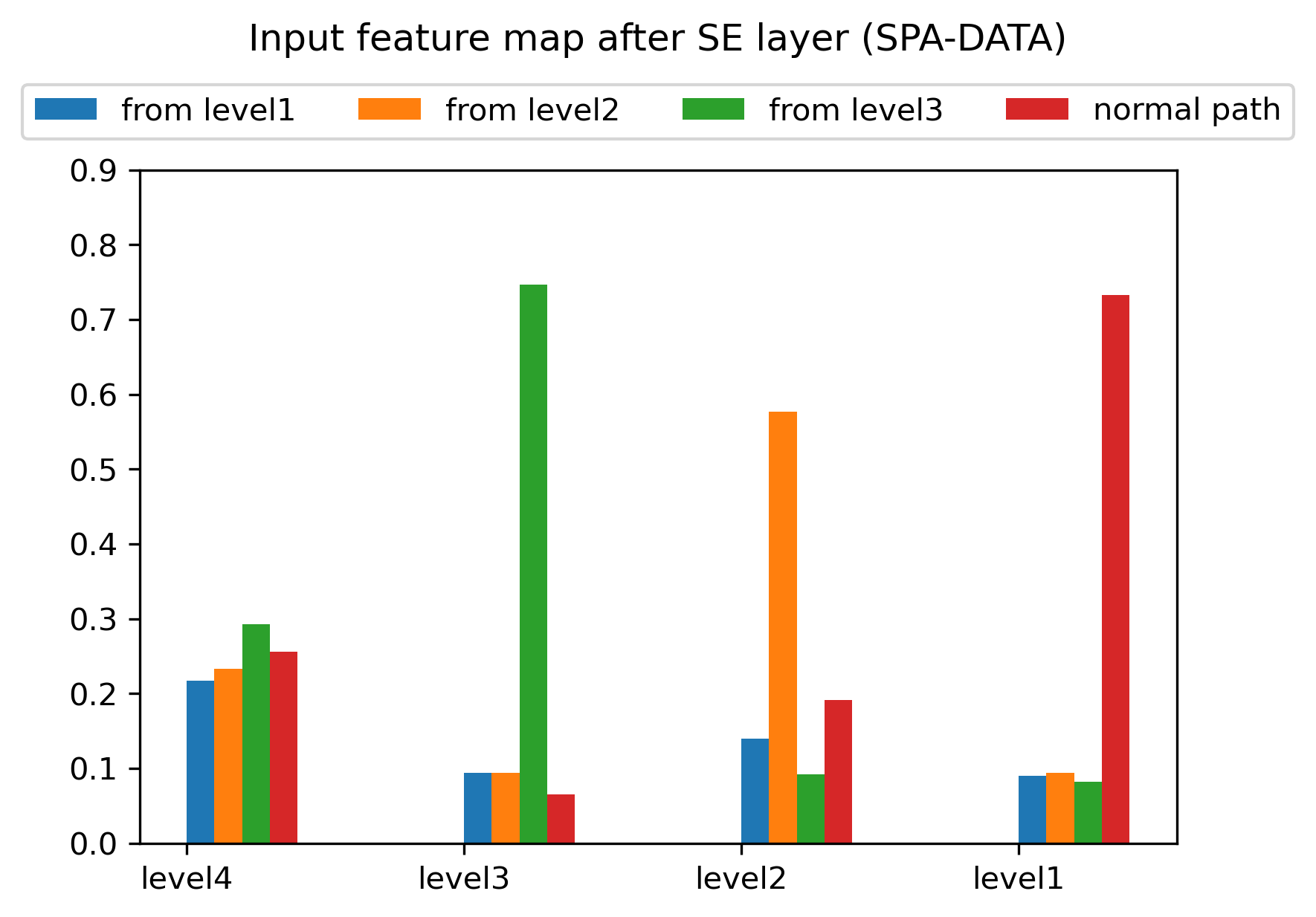}} \\
		\multicolumn{3}{c}{(c)} &
		\multicolumn{3}{c}{(d)}
		
	\end{tabular}
	\caption{ Intensity analysis of channel-wise attentions at each MLC on Rain200H and SPA-DATA datasets.}

	\label{fig:se_analysis}
\end{figure*}

\subsection{Analysis on multi-level features}

To achieve insight as to how much each level contributes in deraining process for each  connection, we measure the feature importance of channel-wise attention in SE layer at each connection.We measure the feature importance as follows: \\
As presented in Section \ref{sec:multi-level connections}, the features of each level in the encoder part are aggregated at level $l$ in the decoder part, 
\begin{align}
    D_{concat}^l &=  (\bigoplus_{i=1}^{3} H_{i}^{l} ( E_{out}^{i} ) )\oplus H_{up}(D_{out}^{l+1}) \\ 
     &=  \tilde{D}_{1}^l \oplus \tilde{D}_{2}^l \oplus \tilde{D}_{3}^l \oplus \tilde{D}_{normal}^l,
\end{align} 
% where $\tilde{D}_{i}^l = H_{i}^{l} ( E_{out}^{i} ) $ and $ \tilde{D}_{normal}^l= H_{up}(D_{out}^{l+1})  $ .}
where $\tilde{D}_{i}^l$ and $\tilde{D}_{normal}^l$ are the results of $H_{i}^{l} ( E_{out}^{i} ) $ and $ H_{up}(D_{out}^{l+1}) $, respectively. Note that $D_{out}^{l+1}$ means the output of previous layer. Afterwards, $D_{concat}^l$ is fed into SE layer $f_{SE}$ as

\begin{align}
    f_{SE}(D_{concat}^l) &=  f_{SE}( [\tilde{D}_{1}^l , \tilde{D}_{2}^l, \tilde{D}_{3}^l , \tilde{D}_{normal}^l] ) \\
    &= [\hat{D}_{1}^l , \hat{D}_{2}^l, \hat{D}_{3}^l , \hat{D}_{normal}^l], 
\end{align} 
where, each $\hat{D}_{i}^l$ can be considered as corresponding output of $\tilde{D}_{i}^l$ because $f_{SE}$ is channel-wise attention operation. Now, we obtain the feature importance by applying $L_2$ norm and normalization to each of them. 
%  Now, we obtain the feature importance by applying $L^2$ norm and normalization to each feature. 
\begin{align}
    \tilde{\lambda}_i^l &=  \frac{\| \tilde{D}_{i}^l \|_2 }{\sum_j \| \tilde{D}_{j}^l \|_2  } , \; i = 1,2,3,normal  \\ \hat{\lambda}_i^l &=  \frac{\| \hat{D}_{i}^l \|_2 }{\sum_j \| \hat{D}_{j}^l \|_2  }  , \; i = 1,2,3,normal,
\end{align}
% where $\tilde{\lambda}_i^l$ denotes the feature importance before SE layer and $\hat{\lambda}_i^l$ denote the feature importance after SE layer. \\
where $\tilde{\lambda}_i^l$ and $\hat{\lambda}_i^l$ denote the feature importance before and after SE layer, respectively. 

We calculate the feature importance before and after the SE layer for Rain200H and SPA-DATA datasets as described above, and we report the results in Figure \ref{fig:se_analysis}. We find that feature importance is evenly distributed before the SE layer, but more diversely distributed after the SE layer. Combining such results with Table \ref{tab:mlc-ablation} and Figure \ref{fig:mlc_ablation}, we assume that the channel-wise attention guided via SE operation has a crucial contribution to deraining. Furthermore, unspecified distribution of feature importance before the SE layer could cause performance degradation, implying that simple connections such as addition and concatenation could be detrimental to the performance. In this respect, we suggest that the SE layer emphasizes more meaningful features for recovering rainy images at each level.

\section{Conclusion}
In this study, we present the multi-level connections and an adaptive regional attention network structure for single-image deraining. The proposed MCW-Net adaptively aggregates features via connections between multiple levels and the SE block in the background recovery. To utilize rich long-range rain-free background information in the deraining process, we propose a novel WRNL.
The proposed method outperforms existing state-of-the-art methods. In particular, the network restores the details of the input image and almost completely removes rain streaks on both the synthesized and the real-world datasets. Furthermore, additional experiments demonstrate that MCW-Net contributes to other vision tasks by enhancing images degraded under bad weather conditions.

\section*{Acknowledgment}
Myungjoo Kang was supported by the NRF grant

[2021R1A2C3010887] and the ICT R\&D program of
MSIT/IITP [1711117093, 2021-0-00077]

% Use the style of numbering in square brackets.
% If nothing is used, default style will be taken.
% \begin{enumerate}[a)]
% \item 
% \item 
% \item 
% \end{enumerate}  

% \begin{itemize}
% \item 
% \item 
% \item 
% \end{itemize}  

% \begin{description}
% \item[]
% \item[] 
% \item[] 
% \end{description}  

% % % % % % % % % % % % % % % % % % 

% Figure
% \begin{figure}[<options>]
% 	\centering
% 		\includegraphics[<options>]{}
% 	  \caption{}\label{fig1}
% \end{figure}

% \begin{table}[<options>]
% \begin{tabular*}{\tblwidth}{@{}LL@{}}
% \toprule
% \midrule
%  & \\
%  & \\
%  & \\
%  & \\
% \bottomrule
% \end{tabular*}
% \end{table}

% Uncomment and use as the case may be
%\begin{theorem} 
%\end{theorem}

% Uncomment and use as the case may be
%\begin{lemma} 
%\end{lemma}


% \section{}\label{}

\printcredits

\bibliographystyle{cas-model2-names}

\bibliography{cas-refs}

\bio{}
\endbio

\bio{}
\endbio

\end{document}